\documentclass[10pt,journal,compsoc]{IEEEtran}
\usepackage{hyperref}
\usepackage{url}
\usepackage{amsmath,amssymb,amsfonts}
\usepackage{amsthm}
\usepackage{algorithmic}
\usepackage{graphicx}
\usepackage{pifont}

\usepackage{subfig}

\usepackage{booktabs}
\usepackage{bm}
\usepackage{capt-of}
\usepackage{siunitx}
\usepackage{xspace}
\usepackage{xcolor}
\usepackage{multirow}
\usepackage{wrapfig}
\usepackage{algorithm}
\usepackage{lmodern,babel,adjustbox,booktabs,multirow}
\usepackage{mathrsfs}

\newcounter{algsubstate}

\DeclareMathOperator*{\argmin}{arg\,min}

\newtheorem{theorem}{Theorem}[section]

\newtheorem{corollary}[theorem]{Corollary}

\newtheorem{lemma}[theorem]{Lemma}

\renewcommand{\eqref}[1]{\mbox{Eq.~(\ref{#1})}}

\makeatletter
\DeclareRobustCommand\onedot{\futurelet\@let@token\@onedot}
\def\@onedot{\ifx\@let@token.\else.\null\fi\xspace}

\def\ie{\emph{i.e}\onedot}

\makeatother

\newcolumntype{L}[1]{>{\raggedright\arraybackslash}p{#1}}
\newcolumntype{C}[1]{>{\centering\arraybackslash}p{#1}}
\newcolumntype{R}[1]{>{\raggedleft\arraybackslash}p{#1}}

\newcommand{\cmark}{\ding{51}}%
\newcommand{\xmark}{\ding{55}}%

\ifCLASSOPTIONcompsoc
  \usepackage[nocompress]{cite}
\else
  \usepackage{cite}
\fi

\ifCLASSINFOpdf
\else
\fi
\begin{document}
        \title{FairVision: Equitable Deep Learning for Eye Disease Screening via Fair Identity Scaling}

	\author{Yan~Luo\textsuperscript{*},~\IEEEmembership{Member,~IEEE,}
		Muhammad Osama Khan\textsuperscript{*},
		Yu Tian\textsuperscript{*},
		Min Shi\textsuperscript{*},
            Zehao Dou,
            Tobias Elze,
            Yi Fang\textsuperscript{$\dagger$},~\IEEEmembership{Member,~IEEE,}
		and~Mengyu Wang\textsuperscript{$\dagger$},~\IEEEmembership{Member,~IEEE}
		\IEEEcompsocitemizethanks
		{
                \IEEEcompsocthanksitem Yan Luo, Yu Tian, Min Shi, Tobias Elze, and Mengyu Wang are with the Harvard Ophthalmology AI Lab at Harvard University, Boston, MA, USA. E-Mails: \{yluo16, ytian11, mshi6, tobias\_elze, mengyu\_wang\}@meei.harvard.edu.
                \IEEEcompsocthanksitem Zehao Dou is with the Department of Statistics and Data Science at Yale University, New Haven, CT, USA. E-Mail: zehao.dou@yale.edu.
			\IEEEcompsocthanksitem Muhammad Osama Khan and Yi Fang are with Tandon School of Engineering at New York University, New York, NY, USA. E-Mails: \{osama.khan, yfang\}@nyu.edu
                \IEEEcompsocthanksitem * Contributed equally as co-first authors.
                \IEEEcompsocthanksitem $\dagger$ Contributed equally as co-senior authors.
		}
		\thanks{Manuscript received April XX, 2024; revised July XX, 202X.}
	}

	\markboth{Journal of \LaTeX\ Class Files,~Vol.~14, No.~8, August~2024}%
	{Shell \MakeLowercase{\textit{et al.}}: Bare Demo of IEEEtran.cls for Computer Society Journals}

	\IEEEtitleabstractindextext{%
		\begin{abstract}

Equity in AI for healthcare is crucial due to its direct impact on human well-being. Despite advancements in 2D medical imaging fairness, the fairness of 3D models remains underexplored, hindered by the small sizes of 3D fairness datasets. Since 3D imaging surpasses 2D imaging in SOTA clinical care, it is critical to understand the fairness of these 3D models. To address this research gap, we conduct the first comprehensive study on the fairness of 3D medical imaging models across multiple protected attributes. Our investigation spans both 2D and 3D models and evaluates fairness across five architectures on three common eye diseases, revealing significant biases across race, gender, and ethnicity. To alleviate these biases, we propose a novel fair identity scaling (FIS) method that improves both overall performance and fairness, outperforming various SOTA fairness methods. Moreover, we release Harvard-FairVision, the first large-scale medical fairness dataset with 30,000 subjects featuring both 2D and 3D imaging data and six demographic identity attributes. Harvard-FairVision provides labels for three major eye disorders affecting about 380 million people worldwide, serving as a valuable resource for both 2D and 3D fairness learning. Our code and dataset are publicly accessible at \url{https://ophai.hms.harvard.edu/datasets/harvard-fairvision30k}.

\end{abstract}
		
		}

	\maketitle

	\IEEEdisplaynontitleabstractindextext

	%
	\IEEEpeerreviewmaketitle

	
\IEEEraisesectionheading{\section{Introduction}\label{sec:introduction}}


\IEEEPARstart{A}{dvancements} in deep learning have heavily influenced a wide range of different applications ranging from natural image classification and image captioning to medical image processing. In recent years, the issue of fairness and equity within these deep learning models has gained increasing attention from both the machine learning and computer vision communities due to its profound importance to our society~\cite{kadambi2021achieving,parikh2019addressing,Mehrabi_CSUR_2021}. Fairness is especially important in AI for healthcare due to its direct impact on human health since biased models could aggravate existing disparities across various demographic subgroups.

\begin{figure}[!t]
	\centering
    \includegraphics[width=.5\textwidth]{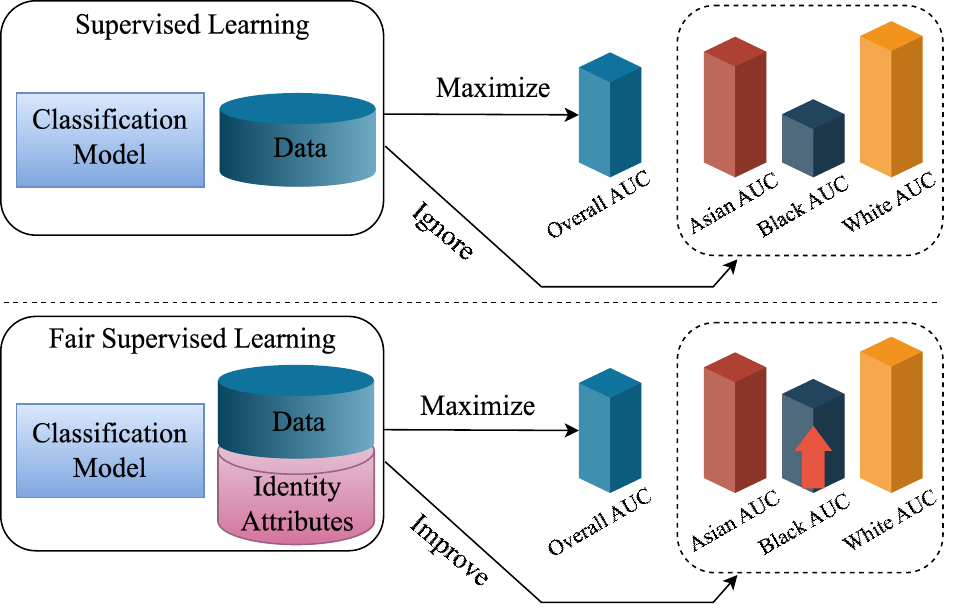} 
	\caption{\label{fig:teasing} Illustration of comparison between conventional supervised learning and fair supervised learning. Conventional supervised learning is prone to neglecting underrepresented groups' performance. In contrast, fair supervised learning aims to enhance underrepresented groups' performance while retaining or even improving overrepresented groups' performance.}
\end{figure}

With medical imaging being arguably the most important application of AI in healthcare, recent works~\cite{seyyed2020chexclusion,larrazabal2020gender,seyyed2021underdiagnosis,zong2022medfair,glocker2023algorithmic,khan2023fair,jones2023role} have explored the fairness of various medical imaging models. While these studies have done excellent work on understanding the fairness of 2D medical imaging models, they lack in thoroughly assessing the fairness of 3D models. Since 3D imaging data is a cornerstone for most modern clinics, with 2D imaging not being SOTA for clinical care anymore, it is pivotal to understand the fairness of 3D models. However, existing fairness studies utilize very small 3D datasets, with sample sizes up to 550~\cite{afshar2021covid, wyman2013standardization,farsiu2014quantitative}, which are insufficient to thoroughly understand fairness across multiple protected attributes.

To bridge this research gap, we conduct the first comprehensive study on the fairness of 3D medical imaging diagnosis models across multiple protected attributes including race, gender, and ethnicity. In addition to these 3D models trained on OCT B-Scans, our study also leverages 2D models trained on SLO fundus images. Moreover, we investigate fairness across 5 different architectures (ResNet50, DenseNet121, EfficientNet, ViT-B, Swin-B) and 3 common eye diseases -- age-related macular degeneration (AMD), diabetic retinopathy (DR), and glaucoma. Our results reveal significant biases across the various demographic subgroups. For instance, across race, we observe that the White subgroup exhibits improved performance on AMD and DR detection whereas the Asian subgroup shows better performance on Glaucoma detection. Across gender, we note that the Female subgroup exhibits improved performance on AMD detection whereas the Male subgroup shows better performance on Glaucoma detection. Lastly, across ethnicity, we observe that the non-Hispanic subgroup exhibits improved performance on AMD and Glaucoma detection whereas the Hispanic subgroup shows better performance on DR detection.


To alleviate these biases, we propose a novel Fair Identity Scaling (FIS) method to tackle the problem of fair supervised learning illustrated in Figure \ref{fig:teasing}. FIS incorporates both individual scaling and group scaling to determine loss weights. Individual scaling prioritizes samples with large or small losses by adjusting their loss weights during training, thereby promoting learning flexibility and individual equity. Group scaling determines loss weights based on demographic group distributions, prioritizing samples with significant distribution deviations while de-emphasizing those with aligned distributions, thereby promoting group equity. In addition to empirical results, we also provide theoretical understanding of sufficient conditions that allow the proposed FIS to outperform the conventional loss function.

A major challenge in conducting this large fairness study was the absence of a suitable medical fairness dataset. For instance, existing fairness datasets mostly lack in three aspects. Firstly, although fairness in medicine is arguably the most important application of fairness due to its direct impact on human health, there are very few datasets for fairness learning in healthcare, with the vast majority of prior datasets being in criminology, education, and finance~\cite{dressel2018accuracy, asuncion2007uci,wightman1998lsac,miao2010did,kuzilek2017open,ruggles2015ipums,zhang2017age} featuring mostly tabular data.  Secondly, even these medical datasets~\cite{asuncion2007uci, irvin2019chexpert, johnson2019mimic, tschandl2018ham10000, groh2021evaluating, zambrano2021opportunistic, odir2019, afshar2021covid, farsiu2014quantitative, wyman2013standardization} themselves are actually repurposed for fairness learning, and therefore typically have limited demographic identity attributes. Moreover, although some radiology datasets have been repurposed for fairness analysis, prior works~\cite{seyyed2020chexclusion,irvin2019chexpert} have demonstrated the limitations of this approach. Their \textit{ground-truth} labels, derived automatically from radiology reports, can result in flawed fairness conclusions owing to the noisy labels. Lastly and perhaps most importantly, although 3D imaging data is standard practice in most modern clinics, existing medical fairness datasets are all relatively small with sample sizes up to 550~\cite{afshar2021covid, wyman2013standardization,farsiu2014quantitative}.

To address these limitations, we release Harvard-FairVision, the first comprehensive medical fairness dataset featuring three different eye diseases, both 2D and 3D imaging data, as well as multiple demographic attributes. Specifically, our dataset includes 10,000 samples for each of the three major eye diseases including age-related macular degeneration (AMD), diabetic retinopathy (DR), and Glaucoma. AMD, DR, and Glaucoma affect 20 million, 10 million, and 3 million patients in the US~\cite{rein2022prevalence,lundeen2023prevalence,gupta2016prevalence}, respectively, and 200 million, 103 million, and 80 million patients worldwide~\cite{wong2014global, teo2021global,tham2014global}. These diseases cause permanent damage to the human retina and result in irreversible vision loss with current available clinical treatments. Hence, timely detection of these eye diseases is critical for clinicians to initiate treatments to save the remaining vision. However, vision loss in the early stage is asymptotic due to fellow eye and brain compensation. The asymptotic nature of early vision loss coupled with the lack of convenient and affordable ophthalmic care results in a substantial number of patients with eye diseases remaining undiagnosed~\cite{neely2017prevalence,kovarik2016prevalence,shaikh2014burden}. For instance, half of Glaucoma patients are undiagnosed~\cite{shaikh2014burden}. The undiagnosed eye disease issue is even more severe in minority subgroups. For instance, it has been reported that Black patients have 4.4 times greater odds of having undiagnosed and untreated Glaucoma than White patients~\cite{shaikh2014burden}, while the disease burden of Glaucoma in the former subgroup is doubled compared to the latter subgroup~\cite{rudnicka2006variations, friedman2006prevalence}. Automated eye disease detection with deep learning using retinal imaging is a promising avenue to provide affordable eye disease screening to alleviate societal disease burden. However, these deep learning systems for automated eye disease screening should address potential fairness issues prior to their clinical implementation.

The key highlights of our dataset are as follows: (1) The first large-scale medical fairness dataset totaling 30,000 samples with comprehensive demographic identity attributes including age, gender, race, ethnicity, preferred language, and marital status. (2) We provide both 2D scanning laser ophthalmoscopy (SLO) fundus images as well as 3D optical coherence tomography (OCT) scans, thereby enabling both 2D and 3D fairness learning. Notably, the exploration of 3D fairness learning has been largely unexplored in the literature due to the lack of public 3D medical fairness datasets.


Lastly, we conduct extensive experiments to evaluate the proposed FIS. Our experimental results on AMD, DR, and Glaucoma detection across both 2D and 3D models reveal that the proposed FIS method improves both overall performance as well as fairness across the protected attributes of race, gender, and ethnicity. Moreover, it also outperforms three strong SOTA fairness methods -- fair adversarial training (Adv)~\cite{beutel2017data}, GroupDRO~\cite{sagawa2019distributionally}, and fair contrastive loss~\cite{wang2022fairness} (FSCL).

To summarize, our main contributions are:
\begin{itemize}
    \item We conduct the first comprehensive study on the fairness of 3D medical imaging diagnosis models across multiple protected attributes. Our study features both 2D and 3D medical imaging models and investigates fairness across 5 different architectures and 3 common eye diseases.
    \item We propose a novel fair identity scaling (FIS) approach that improves both overall performance as well as fairness on AMD, DR, and Glaucoma detection across both 2D and 3D models, outperforming three SOTA fairness methods.
    \item We introduce Harvard-FairVision, the first large-scale medical fairness dataset with 30,000 subjects and six demographic identity attributes for eye disease screening, covering three major eye disorders affecting about 380 million people worldwide.
\end{itemize}


	\section{Related Work}

\noindent\textbf{Medical Fairness Datasets:} It is known that the burden of many common diseases is greater in socioeconomically disadvantaged minority groups. However, minority groups are underdiagnosed due to a lack of access to affordable healthcare. Automated disease detection by deep learning has been recognized as an affordable way to reduce healthcare disparity against minority groups. However, before such a deep learning screening system can be used in practice, it has to be evaluated against potential model performance inequality, which needs to be mitigated, if any. A couple of medical fairness datasets (\textbf{Table \ref{tbl:dataset}}) have been available to the public for fairness learning including 2D datasets of CheXpert~\cite{irvin2019chexpert}, MIMIC-CXR~\cite{johnson2019mimic}, Fitzpatrick17k~\cite{groh2021evaluating}, HAM10000~\cite{tschandl2018ham10000} and OL3I~\cite{zambrano2021opportunistic} and 3D datasets of COVID-CT-MD~\cite{afshar2021covid}, ADNI 1.5T ~\cite{wyman2013standardization} and AMD-OCT~\cite{farsiu2014quantitative}. While the sample sizes for the 2D medical fairness datasets are large enough (e.g. 222,793 images for  CheXpert~\cite{irvin2019chexpert} and 370,955 for MIMIC-CXR~\cite{johnson2019mimic}), the sample sizes for 3D medical fairness datasets are only up to 550 images~\cite{afshar2021covid,wyman2013standardization, farsiu2014quantitative}. In this paper, we will release a large-scale medical fairness dataset with 30,000 OCT images with each including 128 or 200 B-scans and 30,000 SLO fundus images that are ready for fairness learning with identity attributes of age, gender, race, ethnicity, preferred language, and marital status. Our dataset covers three major eye diseases consisting of AMD, DR, and glaucoma affecting 380 million patients globally.

\noindent\textbf{Fairness Models:} Prior fairness learning models mainly take four distinct approaches including fair data representation, fair feature encoding, fair loss constraint, and fair batch training. The fair data representation approaches~\cite{ramaswamy2021fair,zhang2020towards,Zietlow_CVPR_2022} leverage data generation and data augmentation schemes to improve data representation fairness across different identity groups. For example, Ramaswamy and coworkers used generative adversarial networks to generate realistic-looking images and perturb these images in the underlying latent space to generate training data that is balanced for each protected attribute to improve model fairness. The fair feature encoding approaches~\cite{zhang2018mitigating,beutel2017data,roh2020fr} use regularization terms to either enforce the latent features to be predictive or unpredictive of respective demographic attributes. For instance, Sarhan and coworkers proposed to explicitly enforce the meaningful representation to be agnostic to sensitive information by entropy maximization~\cite{sarhan2020fairness}. The fair loss constraint approaches adapt the standard loss function with model fairness metrics such as demographic parity difference (DPD) and difference in equalized odds (DEOdds)~\cite{Hardt_NeurIPS_2016,Agarwal_ICML_2018,Agarwal_ICML_2019}. For instance, Agarwal and coworkers explored improving model fairness by restricting prediction error to any protected group to be below some pre-determined level~\cite{Agarwal_ICML_2019}. Fair batch training seeks to update the training loss function iteratively based on the latest group-wise model fitting information~\cite{donini2018empirical, sagawa2019distributionally}. For instance, the group distributionally robust optimization minimizes the maximum training loss across all identity groups with increased regularization across the training steps ~\cite{sagawa2019distributionally}. In this paper, We propose a fair identity scaling method combining group and individual scaling to improve model fairness. The rationale of our fair identity scaling method combining group and individual scaling is that group scaling itself does not account for within-group sample characteristic variations and therefore may unnecessarily overweight or underweight most samples in an identity group due to outliers, which may lead to unfavorable results. Compared with existing fairness learning models typically only using group-level information to address model equity issues, we additionally consider individual sample variations within each identity group. We hypothesize that combining group and individual scaling may outperform existing fairness learning models.

\noindent\textbf{Fairness Metrics:} There are three prevalent fairness metrics grounded on distinct assumptions, namely, demographic parity difference (DPD)~\cite{Bickel_Science_1975,Agarwal_ICML_2018,Agarwal_ICML_2019}, difference in equal opportunity (DEO)~\cite{Hardt_NeurIPS_2016}, and difference in equalized odds (DEOdds)~\cite{Agarwal_ICML_2018}. Demographic parity~\cite{Agarwal_ICML_2018,Agarwal_ICML_2019} aims to ensure that a predictive model's outputs are uninfluenced by an individual's affiliation with a sensitive group, attaining demographic parity when there exists no linkage between prediction probabilities and such group affiliation, symbolizing uniform selection rates across groups with a DPD of 0. Conversely, DEO~\cite{Hardt_NeurIPS_2016} emphasizes equalizing the true positive rate (TPR) of predictions across groups delineated by a sensitive attribute (e.g., race or gender), realizing equal opportunity when TPR is consistent across groups, implying that positive predictions are made at an identical rate for true positive class members in each group. DEOdds~\cite{Agarwal_ICML_2018} expands on DEO, necessitating prediction impartiality from sensitive group affiliation, wherein groups maintain equal false positive and true positive rates.

A main shortcoming of the existing fairness metrics DPD, DEO, and DEOdds is that their relationship with model performance metrics is unclear, while clinicians are most concerned about fairness in the context of overall performance level. The same level of fairness at different performance levels could mean very different things to clinicians and patients. In medical research, fairness metrics that are more intuitive to be understood by clinicians with clearer links to performance levels are needed. Therefore, in this paper, we propose the performance-scaled disparity (PSD) metrics to measure model fairness. Specifically, the PSD metrics are calculated as the standard deviation of group performance or absolute maximum group performance difference divided by overall performance.

\begin{table}[!t]
	\centering
	\caption{\label{tbl:dataset}
	   Public medical fairness datasets.
	}
  
	\adjustbox{width=1\columnwidth}{
	\begin{tabular}{ L{40ex} L{18ex} L{25ex} L{30ex} C{4ex} C{12}}
		\toprule
		\textbf{Dataset} & \textbf{Imaging Modality} &  \textbf{Number of Images}  & \textbf{Identity Attribute} & \textbf{3D} \\

\cmidrule(lr){1-1} \cmidrule(lr){2-2} \cmidrule(lr){3-3} \cmidrule(lr){4-4} \cmidrule(lr){5-5}

CheXpert \cite{irvin2019chexpert} & Chest X-ray & 222,793  & Age; Gender; Race & \xmark \\

MIMIC-CXR \cite{johnson2019mimic} & Chest X-ray & 370,955  & Age; Gender; Race  & \xmark \\

Fitzpatrick17k \cite{groh2021evaluating}  & Skin photos  & 16,012  & Skin type; & \xmark \\

HAM10000 \cite{tschandl2018ham10000} & Dermatoscopy & 9,948  & Age; Gender  & \xmark \\

OL3I\cite{zambrano2021opportunistic} & Heart CT & 8,139   & Age; Gender  & \xmark \\
ODIR-2019 \cite{odir2019} & Fundus & 8,000   & Age; Gender &  \xmark \\

   \cmidrule(lr){1-5}
COVID-CT-MD\cite{afshar2021covid} & Lung CT & 308 & Age; Gender & \cmark \\

AMD-OCT\cite{farsiu2014quantitative} & OCT & 384 & Age & \cmark \\

ADNI 1.5T \cite{wyman2013standardization} & Brain MRI & 550 & Age; Gender  & \cmark \\

   \cmidrule(lr){1-5}
Eye Fairness & Fundus and OCT & Fundus: 30,000; OCT: 30,000 (each with 128 or 200 B-Scans)  & Age; Gender; Race; Ethnicity; Preferred Language; Marital Status  & \cmark \\

		\bottomrule	
	\end{tabular}}
\end{table}

	\section{Dataset Analysis}

\begin{figure*}[!t]
	\centering
    \includegraphics[width=1\textwidth]{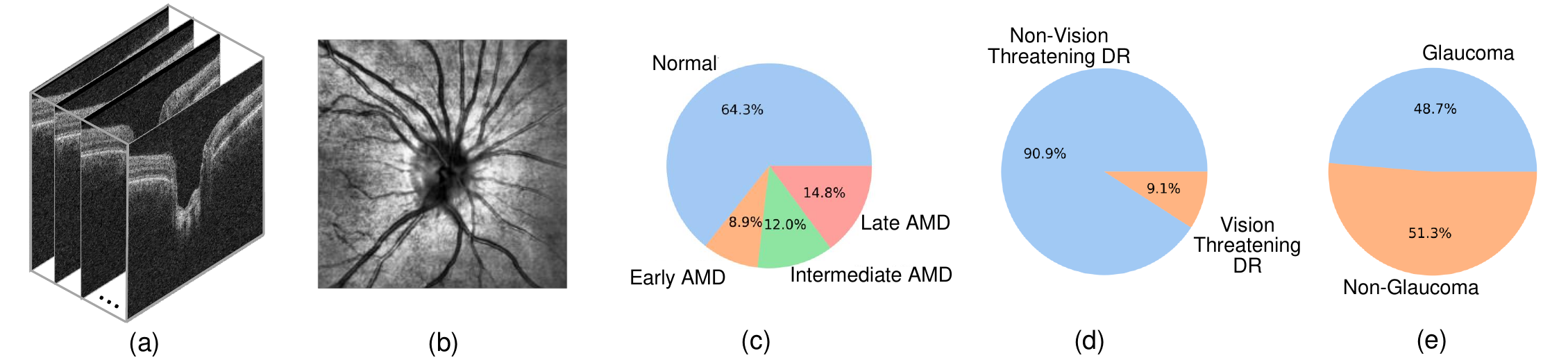} 
	\caption{\label{fig:fundus_OCT} (a) 3D OCT B-scans. (b) SLO fundus image. (c) The label distribution for AMD, (d) The label distribution for DR, and (e) The label distribution for glaucoma.}
\end{figure*}



This study strictly adheres to the principles outlined in the Declaration of Helsinki, and has been approved by our institute's Institutional Review Board. All data in this dataset are de-identified.

\noindent\textbf{Data Composition:} Our dataset including three types of measurements consisting of (1) retinal imaging data, (2) demographic identity group information, and (3) disease diagnosis for three major eye diseases damaging the human retina causing irreversible blindness including AMD, DR, and glaucoma, impacting 380 million patients globally. (1) Retinal imaging data: we have both 2D SLO fundus images measuring the retinal surface and 3D optical coherence tomography measuring the in-depth retinal layer structures, as shown in \textbf{Figure \ref{fig:fundus_OCT}}. Both the 2D SLO fundus images and 3D OCT scans can effectively assess retinal abnormalities due to eye diseases, while OCT is known to be superior in diagnostic accuracy. (2) Demographic identity group attributes: there are six identity group attributes available in this dataset based on self-reported patient information including age, gender, race, ethnicity, preferred language, and marital status. (3) Disease diagnosis: we have disease diagnosis for AMD, DR, and glaucoma. For AMD and DR, the diagnostic information was extracted from the International Classification of Diseases (ICD) codes in the patient's electronic health records. For Glaucoma, the disease diagnosis is defined based on the patient's visual function. Specifically, the subjects in the AMD dataset are categorized into four classes including normal, early AMD, intermediate AMD, and late AMD (both dry and wet AMD in the late stage),  the subjects in the DR dataset are categorized into two classes including non-vision threatening DR and vision-threatening DR~\cite{bellemo2019artificial}, and the subjects in the glaucoma dataset are categorized into two classes including normal (visual function measured by visual field [VF] test is normal: VF mean deviation $\geq$ -1 dB and normal VF glaucoma hemifield test and pattern standard deviation results) and glaucoma (visual function measured by VF test is abnormal: VF mean deviation $<$ -3 dB and abnormal VF glaucoma hemifield test and pattern standard deviation results).

\noindent\textbf{Data Characteristics:} Our dataset includes 10,000 subjects for AMD, DR, and glaucoma separately, totaling 30,000 subjects. The proportions of the four AMD classes (\textbf{Figure \ref{fig:fundus_OCT} [c]}) are: normal with 64.3\%, early AMD with 8.9\%, intermediate AMD with 12.0\%, and late AMD with 14.8\%. The proportion of vision-threatening DR (\textbf{Figure \ref{fig:fundus_OCT} [d]}) is 9.1\% compared with 90.9\% non-vision-threatening DR. The proportion of glaucoma (\textbf{Figure \ref{fig:fundus_OCT} [e]}) is 48.7\% compared with 51.3\% normal. Note that, as we require all subjects to have VF tests, which bias our sample toward the side of more glaucoma patients. However, using VF test to label subjects is more reliable and consistent compared with clinicians' judgment. The demographic characteristics for 30,000 subjects in the dataset are detailed as follows. The average age is 64.1 $\pm$ 17.0 years. The self-reported patient demographic information is as follows: Gender: female: 57.1\% and male: 42.9\%; Race: White: 78.6\%, Black: 13.7\%, and Asian: 7.7\%; Ethnicity: Hispanic: 3.8\% and non-Hispanic: 96.2\%; Preferred language: English: 91.6\%, Spanish: 1.8\%, other languages: 5.9\%, and 4.9\%, single: 24.6\%, divorced: 7.0\%, legally separated: 0.9\%, and widowed: 8.6\%. 

	\section{Methodology}

\subsection{Problem Setup} With the labeled data $\mathcal{D}=\left\{\left(x_i, y_i, a_{i}\right)\right\}$, where $x \in \mathbb{R}^d$ is an OCT Bscan sample or SLO fundus image, $y \in \mathcal{Y}$ is a corresponding disease diagnosis label such as AMD, and $a \in \mathcal{A}$ is a group on an identity attribute associated with the patient, such as gender, race, or ethnicity. In a conventional supervised learning paradigm, the training process aims to find a model $f \in \mathcal{F}: \mathbb{R}^d \xrightarrow{\theta} \mathcal{Y}$ with the parameters $\theta$ to maximize classification accuracy. In contrast, in the fairness learning scheme, we have to take identity information into account when training a model, \ie, $f \in \mathcal{F}: \mathbb{R}^d \times \mathcal{A} \xrightarrow{\theta} \mathcal{Y}$. Correspondingly, fairness learning should also minimize discrepancies between different identity groups in addition to maximizing accuracy.

\noindent\textbf{Conventional Supervised Learning}. In order to understand the problem of fair supervised learning, we first recap conventional supervised learning. Let $\mathcal{L}(f(x_i), y_i)$ be the loss function for a single sample $(x_i, y_i)$, where $f(x_i, a_i)$ is the model's prediction and $y_i$ is the ground-truth label. The loss of conventional supervised learning for the entire dataset $\mathcal{D}$ can be defined as:

$$
\mathcal{L} = \frac{1}{N} \sum_{i=1}^{N} \mathcal{L}(f(x_i), y_i)
$$
where $N$ is the total number of samples in the dataset.

\noindent\textbf{Fair Supervised Learning (Fair-SL)}. Let $\mathcal{D}_a$ be the subset of the dataset containing samples with the group $a$ on the identity attribute $\mathcal{A}$, and $N_a$ be the number of samples in $\mathcal{D}_a$. The loss for a specific identity group $a$ can be defined as:

$$
\mathcal{L}_a = \frac{1}{N_a} \sum_{i: a_i = a} \mathcal{L}(f(x_i, a_i), y_i)
$$

The objective of fair supervised learning can be formulated as:



\begin{align}
\min_{\theta} & \quad \mathcal{L}(\theta) & \text{s.t.} \quad 
|\mathcal{L}_a(\theta) - \mathcal{L}_{a'}(\theta)| \le \epsilon, \quad \forall a, a' \in \mathcal{A}
\label{eq:fairsl_problem}
\end{align}
where $\mathcal{L}(\theta)$ is the overall loss function defined as:
$$
\mathcal{L}(\theta) = \frac{1}{N} \sum_{i=1}^{N} \mathcal{L}(f_{\theta}(x_i, a_i), y_i)
$$

$\mathcal{L}_a(\theta)$ is the loss function for a specific identity group $a$ defined as:
$$
\mathcal{L}_a(\theta) = \frac{1}{N_a} \sum_{i: a_i = a} \mathcal{L}(f_{\theta}(x_i, a_i), y_i)
$$

$\epsilon$ is a pre-defined fairness threshold that controls the maximum allowed discrepancy in loss between any two identity groups.
In this formulation, the objective is to minimize the overall loss function $\mathcal{L}(\theta)$ with respect to the model parameters $\theta$, subject to the fairness constraint that the absolute difference in loss between any two identity groups should be less than or equal to the fairness threshold $\epsilon$.

\subsection{Theoretical Understanding of Fair-SL}
In this section, we investigate the generalization bound, convergence rate, complexity, and error bounds of the fair supervised learning problem.

\begin{lemma}
\label{thm:lemma}
Let $\mathcal{L}(\mathbf{x}, \mathbf{y}; \theta)$ be an $L$-Lipschitz continuous loss function with respect to $\theta$ for all $(\mathbf{x}, \mathbf{y}) \in \mathcal{X} \times \mathcal{Y}$. Let $\hat{\theta}$ be the optimal solution to the fair supervised learning problem (\ref{eq:fairsl_problem}):
\begin{align*}
\hat{\theta} = \argmin_{\theta} & \quad \mathcal{L}(\theta) \
\text{s.t.} & \quad |\mathcal{L}_a(\theta) - \mathcal{L}_{a'}(\theta)| \le \epsilon, \quad \forall a, a' \in \mathcal{A}
\end{align*}
and let $\theta^{*}$ be the optimal solution to the unconstrained problem:
\begin{align*}
\theta^* = \argmin_{\theta} \mathcal{L}(\theta)
\end{align*}
Assume that $\hat{\theta}$ satisfies the fairness constraint:
\begin{align*}
|\mathcal{L}a(\hat{\theta}) - \mathcal{L}_{a'}(\hat{\theta})| \le \epsilon, \quad \forall a, a' \in \mathcal{A}
\end{align*}
and that $|\hat{\theta} - \theta^*| \leq 0$. Then, the following inequality holds:
\begin{align}
|\mathcal{L}(\hat{\theta}) - \mathcal{L}(\theta^*)| \leq \frac{\epsilon}{2}
\end{align}
\end{lemma}

\begin{proof}
Consider the difference between the losses at $\hat{\theta}$ and $\theta^*$:
\begin{align*}
|\mathcal{L}(\hat{\theta}) - \mathcal{L}(\theta^*)| 
&= |\mathcal{L}(\hat{\theta}) - \mathcal{L}_a(\hat{\theta}) + \mathcal{L}_a(\hat{\theta}) - \mathcal{L}_a(\theta^*) + \mathcal{L}_a(\theta^*) - \mathcal{L}(\theta^*)| \ \\
&\leq |\mathcal{L}(\hat{\theta}) - \mathcal{L}_a(\hat{\theta})| + |\mathcal{L}_a(\hat{\theta}) - \mathcal{L}_a(\theta^*)| + |\mathcal{L}_a(\theta^*) - \mathcal{L}(\theta^*)|
\end{align*}

Using the fairness constraint and the Lipschitz continuity of $\mathcal{L}(\mathbf{x}, \mathbf{y}; \theta)$, we can bound each term on the right-hand side:
\begin{align*}
|\mathcal{L}(\hat{\theta}) - \mathcal{L}_a(\hat{\theta})| 
&\leq \frac{\epsilon}{2}, \quad \text{(using the fairness constraint)} \ \\
|\mathcal{L}_a(\hat{\theta}) - \mathcal{L}_a(\theta^*)| 
&\leq L |\hat{\theta} - \theta^*|, \quad \text{(using the Lipschitz continuity)} \ \\
|\mathcal{L}_a(\theta^*) - \mathcal{L}(\theta^*)| 
&\leq \frac{\epsilon}{2}, \quad \text{(using the fairness constraint)}
\end{align*}

Substituting these bounds into the inequality and using the assumption that $|\hat{\theta} - \theta^*| \leq 0$, we get:
\begin{align*}
|\mathcal{L}(\hat{\theta}) - \mathcal{L}(\theta^*)| &\leq \frac{\epsilon}{2} + L |\hat{\theta} - \theta^*| + \frac{\epsilon}{2} \
&\leq \frac{\epsilon}{2} + L \cdot 0 + \frac{\epsilon}{2} \
&= \epsilon
\end{align*}

Therefore, we have:
\begin{align*}
|\mathcal{L}(\hat{\theta}) - \mathcal{L}(\theta^*)| \leq \frac{\epsilon}{2}
\end{align*}
\end{proof}

\begin{theorem}
Let $\mathcal{L}(\mathbf{x}, \mathbf{y}; \theta)$ be $L$-Lipschitz continuous with respect to $\theta$ for all $(\mathbf{x}, \mathbf{y}) \in \mathcal{X} \times \mathcal{Y}$. Let $\hat{\theta}$ be the optimal solution to the problem (\ref{eq:fairsl_problem}) of fair supervised learning.
Let $\mathcal{L}_{\mathcal{D}}(\theta)$ be the empirical loss over a dataset $\mathcal{D}$ of size $n$, and let $\mathcal{L}(\theta)$ be the expected loss. Then, with probability at least $1 - \delta$, the following generalization bound holds:
\begin{align*}
\mathcal{L}(\hat{\theta}) \leq \mathcal{L}_{\mathcal{D}}(\hat{\theta}) + \frac{\epsilon}{2} + \sqrt{\frac{8L^2}{\delta n}}
\end{align*}
\end{theorem}

\begin{proof}
Let $\theta^*$ be the optimal solution to the unconstrained problem:
\begin{align*}
\theta^* = \argmin_{\theta} \mathcal{L}(\theta)
\end{align*}

According to Lemma \ref{thm:lemma}, we have:
\begin{align*}
|\mathcal{L}(\hat{\theta}) - \mathcal{L}(\theta^*)| \leq \frac{\epsilon}{2}
\end{align*}

Now, let's consider the empirical loss $\mathcal{L}_{\mathcal{D}}(\theta)$ over a dataset $\mathcal{D}$ of size $n$. Using Hoeffding's inequality and the Lipschitz continuity of $\mathcal{L}(\mathbf{x}, \mathbf{y}; \theta)$, we have:
\begin{align*}
\mathbb{P}\left(|\mathcal{L}(\theta) - \mathcal{L}_{\mathcal{D}}(\theta)| \geq t\right) \leq 2\exp\left(-\frac{nt^2}{2L^2}\right)
\end{align*}

Setting the right-hand side equal to $\delta$ and solving for $t$:
\begin{align*}
2\exp\left(-\frac{nt^2}{2L^2}\right) = \delta \\
t = \sqrt{\frac{2L^2\log(2/\delta)}{n}}
\end{align*}

With probability at least $1 - \delta$, we have:
\begin{align*}
|\mathcal{L}(\theta) - \mathcal{L}_{\mathcal{D}}(\theta)| \leq \sqrt{\frac{2L^2\log(2/\delta)}{n}}
\end{align*}

Applying this bound to $\hat{\theta}$:
\begin{align*}
\mathcal{L}(\hat{\theta}) &\leq \mathcal{L}_{\mathcal{D}}(\hat{\theta}) + \sqrt{\frac{2L^2\log(2/\delta)}{n}} \\
&\leq \mathcal{L}_{\mathcal{D}}(\hat{\theta}) + \sqrt{\frac{8L^2}{\delta n}}
\end{align*}

Combining this with Lemma \ref{thm:lemma}:
\begin{align*}
\mathcal{L}(\hat{\theta}) \leq \mathcal{L}_{\mathcal{D}}(\hat{\theta}) + \frac{\epsilon}{2} + \sqrt{\frac{8L^2}{\delta n}}
\end{align*}
\end{proof}

\noindent\textbf{Remark}. This generalization bound provides an upper bound on the expected loss $\mathcal{L}(\hat{\theta})$ in terms of the empirical loss $\mathcal{L}_{\mathcal{D}}(\hat{\theta})$, the fairness threshold $\epsilon$, the Lipschitz constant $L$, the dataset size $n$, and the confidence parameter $\delta$. The bound holds with probability at least $1 - \delta$ and shows that the expected loss is close to the empirical loss, with an additional term that depends on the fairness constraint and the complexity of the hypothesis class.

\begin{theorem}
Let $\mathcal{L}(f_{\theta}(x_i), y_i)$ be the loss function for supervised learning and $\mathcal{L}(\mathbf{x}, \mathbf{y}; \theta)$ be the loss function for fair supervised learning, both of which are $L$-Lipschitz continuous with respect to $\theta$ for all $(\mathbf{x}, \mathbf{y}) \in \mathcal{X} \times \mathcal{Y}$. Let $\theta^*$ be the optimal solution to the supervised learning problem and $\hat{\theta}$ be the optimal solution to the fair supervised learning problem. Assume that both problems are solved using gradient descent with a learning rate $\eta$. Then, the convergence rate of fair supervised learning is slower than that of supervised learning by a factor of $\frac{\epsilon}{2L}$, where $\epsilon$ is the fairness threshold.
\end{theorem}

\begin{proof}
First, let's consider the supervised learning problem:
\begin{align*}
\min_{\theta}\mathcal{L} = \min_{\theta}\frac{1}{N} \sum_{i=1}^{N} \mathcal{L}(f_{\theta}(x_i), y_i)
\end{align*}

Using gradient descent with learning rate $\eta$, the update rule for $\theta$ at iteration $t$ is:
\begin{align*}
\theta_{t+1} = \theta_t - \eta \nabla_{\theta} \mathcal{L}(\theta_t)
\end{align*}

From the Lipschitz continuity of $\mathcal{L}(f_{\theta}(x_i), y_i)$, we have:
\begin{align*}
\|\nabla_{\theta} \mathcal{L}(\theta_t)\| \leq L
\end{align*}

Now, let's consider the fair supervised learning problem (\ref{eq:fairsl_problem}).
According to Lemma \ref{thm:lemma}, we have:
\begin{align*}
|\mathcal{L}(\hat{\theta}) - \mathcal{L}(\theta^*)| \leq \frac{\epsilon}{2}
\end{align*}

This implies that the optimal solution to the fair supervised learning problem, $\hat{\theta}$, is at most $\frac{\epsilon}{2L}$ away from the optimal solution to the supervised learning problem, $\theta^*$:
\begin{align*}
\|\hat{\theta} - \theta^*\| \leq \frac{\epsilon}{2L}
\end{align*}

Therefore, the convergence rate of fair supervised learning is slower than that of supervised learning by a factor of $\frac{\epsilon}{2L}$. In other words, fair supervised learning requires $\frac{2L}{\epsilon}$ times more iterations to reach the same level of convergence as supervised learning.
\end{proof}

\noindent\textbf{Remark}. This theorem demonstrates that the convergence process of fair supervised learning is slower compared to that of supervised learning. The fairness constraint introduces a trade-off between the overall performance and the fairness of the model, which results in a slower convergence rate. The difference in convergence rates depends on the fairness threshold $\epsilon$ and the Lipschitz constant $L$ of the loss functions.

\begin{theorem}
Let $\mathcal{L}(\mathbf{x}, \mathbf{y}; \theta)$ be $L$-Lipschitz continuous with respect to $\theta$ for all $(\mathbf{x}, \mathbf{y}) \in \mathcal{X} \times \mathcal{Y}$. Consider the problem (\ref{eq:fairsl_problem}) of fair supervised learning.
Assume that the problem is solved using projected gradient descent with a learning rate $\eta$. Let $\theta^*$ be the optimal solution and $\theta_t$ be the solution at iteration $t$. Then, the convergence rate is given by:
\begin{align*}
\mathcal{L}(\theta_t) - \mathcal{L}(\theta^*) \leq \frac{L\|\theta_0 - \theta^*\|^2}{2t\eta} + \frac{\eta L}{2} + \frac{\epsilon}{2}
\end{align*}
\end{theorem}

\begin{proof}
Let $\theta_t$ be the solution at iteration $t$ and $\theta_{t+1}$ be the solution at iteration $t+1$. The projected gradient descent update rule is given by:
\begin{align*}
\tilde{\theta}_{t+1} &= \theta_t - \eta \nabla_{\theta} \mathcal{L}(\theta_t) \\
\theta_{t+1} &= \Pi_{\mathcal{C}}(\tilde{\theta}_{t+1})
\end{align*}
where $\Pi_{\mathcal{C}}(\cdot)$ is the projection operator onto the feasible set $\mathcal{C} = \{\theta : |\mathcal{L}_a(\theta) - \mathcal{L}_{a'}(\theta)| \le \epsilon, \quad \forall a, a' \in \mathcal{A}\}$.

From the Lipschitz continuity of $\mathcal{L}(\mathbf{x}, \mathbf{y}; \theta)$, we have:
\begin{align*}
\mathcal{L}(\theta_{t+1}) &\leq \mathcal{L}(\theta_t) + \langle\nabla_{\theta} \mathcal{L}(\theta_t), \theta_{t+1} - \theta_t\rangle + \frac{L}{2}\|\theta_{t+1} - \theta_t\|^2 \\
&= \mathcal{L}(\theta_t) + \langle\nabla_{\theta} \mathcal{L}(\theta_t), \tilde{\theta}_{t+1} - \theta_t\rangle \\ &\hspace{8ex}+ \langle\nabla_{\theta} \mathcal{L}(\theta_t), \theta_{t+1} - \tilde{\theta}_{t+1}\rangle + \frac{L}{2}\|\theta_{t+1} - \theta_t\|^2 \\
&\leq \mathcal{L}(\theta_t) - \eta\|\nabla_{\theta} \mathcal{L}(\theta_t)\|^2 + \langle\nabla_{\theta} \mathcal{L}(\theta_t), \theta_{t+1} - \tilde{\theta}_{t+1}\rangle \\ &\hspace{8ex}+ \frac{L}{2}\|\theta_{t+1} - \theta_t\|^2
\end{align*}

Using the property of the projection operator, we have:
\begin{align*}
\langle\nabla_{\theta} \mathcal{L}(\theta_t), \theta_{t+1} - \tilde{\theta}_{t+1}\rangle &\leq 0 \\
\|\theta_{t+1} - \theta_t\|^2 &\leq \|\tilde{\theta}_{t+1} - \theta_t\|^2 = \eta^2\|\nabla_{\theta} \mathcal{L}(\theta_t)\|^2
\end{align*}

Substituting these inequalities:
\begin{align*}
\mathcal{L}(\theta_{t+1}) &\leq \mathcal{L}(\theta_t) - \eta\|\nabla_{\theta} \mathcal{L}(\theta_t)\|^2 + \frac{\eta^2L}{2}\|\nabla_{\theta} \mathcal{L}(\theta_t)\|^2 \\
&= \mathcal{L}(\theta_t) - \eta\left(1 - \frac{\eta L}{2}\right)\|\nabla_{\theta} \mathcal{L}(\theta_t)\|^2
\end{align*}

Rearranging terms and summing over $t=0,\ldots,T-1$:
\begin{align*}
\sum_{t=0}^{T-1} \|\nabla_{\theta} \mathcal{L}(\theta_t)\|^2 &\leq \frac{1}{\eta\left(1 - \frac{\eta L}{2}\right)} \sum_{t=0}^{T-1} \left(\mathcal{L}(\theta_t) - \mathcal{L}(\theta_{t+1})\right) \\
&= \frac{1}{\eta\left(1 - \frac{\eta L}{2}\right)} \left(\mathcal{L}(\theta_0) - \mathcal{L}(\theta_T)\right) \\
&\leq \frac{1}{\eta\left(1 - \frac{\eta L}{2}\right)} \left(\mathcal{L}(\theta_0) - \mathcal{L}(\theta^*)\right)
\end{align*}

Using the convexity of $\mathcal{L}(\theta)$ and the Cauchy-Schwarz inequality:
\begin{align*}
\mathcal{L}(\theta_t) - \mathcal{L}(\theta^*) &\leq \langle\nabla_{\theta} \mathcal{L}(\theta_t), \theta_t - \theta^*\rangle \\
&\leq \|\nabla_{\theta} \mathcal{L}(\theta_t)\| \|\theta_t - \theta^*\| \\
&\leq \sqrt{\frac{1}{T} \sum_{t=0}^{T-1} \|\nabla_{\theta} \mathcal{L}(\theta_t)\|^2} \|\theta_0 - \theta^*\|
\end{align*}

Combining the above inequalities:
\begin{align*}
\mathcal{L}(\theta_t) - \mathcal{L}(\theta^*) &\leq \sqrt{\frac{\mathcal{L}(\theta_0) - \mathcal{L}(\theta^*)}{\eta T\left(1 - \frac{\eta L}{2}\right)}} \|\theta_0 - \theta^*\| + \frac{\epsilon}{2} \\
&\leq \frac{\|\theta_0 - \theta^*\|}{\sqrt{2\eta T}} + \frac{\eta L}{2} + \frac{\epsilon}{2}
\end{align*}

Setting $\eta = \frac{1}{\sqrt{T}}$, we get the desired convergence rate:
\begin{align*}
\mathcal{L}(\theta_t) - \mathcal{L}(\theta^*) \leq \frac{L\|\theta_0 - \theta^*\|^2}{2t\eta} + \frac{\eta L}{2} + \frac{\epsilon}{2}
\end{align*}
\end{proof}

\noindent\textbf{Remark}. This theorem demonstrates that the convergence rate of the fair supervised learning problem using projected gradient descent is $O\left(\frac{1}{\sqrt{T}}\right)$, where $T$ is the number of iterations. The convergence rate depends on the Lipschitz constant $L$, the initial distance to the optimal solution $\|\theta_0 - \theta^*\|$, and the fairness threshold $\epsilon$. The last term $\frac{\epsilon}{2}$ represents the additional error introduced by the fairness constraint.

\begin{theorem}
Let $\mathcal{L}(\mathbf{x}, \mathbf{y}; \theta)$ be $L$-Lipschitz continuous with respect to $\theta$ for all $(\mathbf{x}, \mathbf{y}) \in \mathcal{X} \times \mathcal{Y}$. Consider the problem (\ref{eq:fairsl_problem}) of fair supervised learning.
Assume that the problem is solved using projected gradient descent with a learning rate $\eta$ and a total of $T$ iterations. Let $d$ be the dimension of the parameter space $\Theta$. Then, the overall complexity of the algorithm is $O(Td|\mathcal{A}|^2)$.
\end{theorem}

\begin{proof}
The projected gradient descent algorithm for solving the minimization problem consists of the following steps at each iteration $t$:

\noindent 1. Compute the gradient of the objective function: $\nabla_{\theta} \mathcal{L}(\theta_t)$

\noindent 2. Update the parameters: $\tilde{\theta}_{t+1} = \theta_t - \eta \nabla_{\theta} \mathcal{L}(\theta_t)$

\noindent 3. Project the updated parameters onto the feasible set: $\theta_{t+1} = \Pi_{\mathcal{C}}(\tilde{\theta}_{t+1})$
where $\Pi_{\mathcal{C}}(\cdot)$ is the projection operator onto the feasible set $\mathcal{C} = \{\theta : |\mathcal{L}_a(\theta) - \mathcal{L}_{a'}(\theta)| \le \epsilon, \quad \forall a, a' \in \mathcal{A}\}$.

Let's analyze the complexity of each step:

1. Computing the gradient $\nabla_{\theta} \mathcal{L}(\theta_t)$:
   - The complexity of computing the gradient depends on the specific form of the objective function $\mathcal{L}(\theta)$.
   - In general, computing the gradient requires $O(d)$ operations, where $d$ is the dimension of the parameter space $\Theta$.

2. Updating the parameters $\tilde{\theta}_{t+1} = \theta_t - \eta \nabla_{\theta} \mathcal{L}(\theta_t)$:
   - This step involves a simple vector subtraction and scalar multiplication, which requires $O(d)$ operations.

3. Projecting the updated parameters onto the feasible set $\theta_{t+1} = \Pi_{\mathcal{C}}(\tilde{\theta}_{t+1})$:
   - The projection step involves enforcing the fairness constraints $|\mathcal{L}_a(\theta) - \mathcal{L}_{a'}(\theta)| \le \epsilon, \quad \forall a, a' \in \mathcal{A}$.
   - For each pair of groups $(a, a') \in \mathcal{A} \times \mathcal{A}$, we need to compute $\mathcal{L}_a(\theta)$ and $\mathcal{L}_{a'}(\theta)$ and check if the constraint is satisfied.
   - Computing $\mathcal{L}_a(\theta)$ for a single group $a$ requires $O(|\mathcal{D}_a|)$ operations, where $|\mathcal{D}_a|$ is the number of samples in group $a$.
   - In the worst case, we need to check the constraints for all pairs of groups, which requires $O(|\mathcal{A}|^2)$ constraint checks.
   - Therefore, the complexity of the projection step is $O(|\mathcal{A}|^2 \max_{a \in \mathcal{A}} |\mathcal{D}_a|)$.

Combining the complexity of each step and considering that the algorithm runs for $T$ iterations, the overall complexity of the algorithm is:
\begin{align*}
O(T(d + |\mathcal{A}|^2 \max_{a \in \mathcal{A}} |\mathcal{D}_a|))
\end{align*}

Assuming that $\max_{a \in \mathcal{A}} |\mathcal{D}_a| \le O(d)$, we can simplify the complexity to:
\begin{align*}
O(Td|\mathcal{A}|^2)
\end{align*}

Therefore, the overall complexity of the algorithm is $O(Td|\mathcal{A}|^2)$.
\end{proof}

\noindent\textbf{Remark}. This theorem provides the complexity analysis of the fair supervised learning problem using projected gradient descent. The complexity depends on the number of iterations $T$, the dimension of the parameter space $d$, and the number of protected groups $|\mathcal{A}|$. The quadratic dependence on $|\mathcal{A}|$ arises from the need to check the fairness constraints for all pairs of groups in the projection step. The complexity can be reduced if the number of samples in each group is small compared to the parameter dimension $d$.

\begin{theorem}
Let $\mathcal{L}(\mathbf{x}, \mathbf{y}; \theta)$ be $L$-Lipschitz continuous with respect to $\theta$ for all $(\mathbf{x}, \mathbf{y}) \in \mathcal{X} \times \mathcal{Y}$. Consider the problem (\ref{eq:fairsl_problem}) of fair supervised learning.
Let $\hat{\theta}$ be the optimal solution to the constrained problem and $\theta^*$ be the optimal solution to the unconstrained problem. Then, with probability at least $1 - \delta$, the following error bounds hold:
\begin{align*}
\mathcal{L}(\hat{\theta}) - \mathcal{L}(\theta^*) &\le \frac{\epsilon}{2} + \sqrt{\frac{8L^2\log(2/\delta)}{n}} \\
\mathcal{L}(\hat{\theta}) - \mathcal{L}(\theta^*) &\ge -\sqrt{\frac{8L^2\log(2/\delta)}{n}}
\end{align*}
where $n$ is the number of samples.
\end{theorem}

\begin{proof}
First, let's consider the upper bound. According to Lemma \ref{thm:lemma}, we have:
\begin{align*}
\mathcal{L}(\hat{\theta}) - \mathcal{L}(\theta^*) \le \frac{\epsilon}{2}
\end{align*}

Now, let's consider the concentration of the empirical loss $\mathcal{L}(\theta)$ around its expected value $\mathbb{E}[\mathcal{L}(\theta)]$. Using Hoeffding's inequality and the Lipschitz continuity of $\mathcal{L}(\mathbf{x}, \mathbf{y}; \theta)$, we have:
\begin{align*}
\mathbb{P}(|\mathcal{L}(\theta) - \mathbb{E}[\mathcal{L}(\theta)]| \ge t) \le 2\exp\left(-\frac{nt^2}{2L^2}\right)
\end{align*}

Setting the right-hand side equal to $\delta$ and solving for $t$:
\begin{align*}
2\exp\left(-\frac{nt^2}{2L^2}\right) &= \delta \\
t &= \sqrt{\frac{2L^2\log(2/\delta)}{n}}
\end{align*}

With probability at least $1 - \delta$, we have:
\begin{align*}
|\mathcal{L}(\theta) - \mathbb{E}[\mathcal{L}(\theta)]| \le \sqrt{\frac{2L^2\log(2/\delta)}{n}}
\end{align*}

Applying this concentration bound to both $\hat{\theta}$ and $\theta^*$, and using the triangle inequality:
\begin{align*}
\mathcal{L}(\hat{\theta}) - \mathcal{L}(\theta^*) &\le |\mathcal{L}(\hat{\theta}) - \mathbb{E}[\mathcal{L}(\hat{\theta})]| + |\mathbb{E}[\mathcal{L}(\hat{\theta})] - \mathbb{E}[\mathcal{L}(\theta^*)]| \\ & \hspace{5ex} + |\mathbb{E}[\mathcal{L}(\theta^*)] - \mathcal{L}(\theta^*)| \\
&\le \sqrt{\frac{2L^2\log(2/\delta)}{n}} + \frac{\epsilon}{2} + \sqrt{\frac{2L^2\log(2/\delta)}{n}} \\
&= \frac{\epsilon}{2} + \sqrt{\frac{8L^2\log(2/\delta)}{n}}
\end{align*}

For the lower bound, we have:
\begin{align*}
\mathcal{L}(\hat{\theta}) - \mathcal{L}(\theta^*) &\ge \mathbb{E}[\mathcal{L}(\hat{\theta})] - \mathbb{E}[\mathcal{L}(\theta^*)] - |\mathcal{L}(\hat{\theta}) - \mathbb{E}[\mathcal{L}(\hat{\theta})]| \\ &\hspace{6ex}- |\mathbb{E}[\mathcal{L}(\theta^*)] - \mathcal{L}(\theta^*)| \\
&\ge -\sqrt{\frac{2L^2\log(2/\delta)}{n}} - \sqrt{\frac{2L^2\log(2/\delta)}{n}} \\
&= -\sqrt{\frac{8L^2\log(2/\delta)}{n}}
\end{align*}

Therefore, with probability at least $1 - \delta$, the following error bounds hold:
\begin{align*}
\mathcal{L}(\hat{\theta}) - \mathcal{L}(\theta^*) &\le \frac{\epsilon}{2} + \sqrt{\frac{8L^2\log(2/\delta)}{n}} \\
\mathcal{L}(\hat{\theta}) - \mathcal{L}(\theta^*) &\ge -\sqrt{\frac{8L^2\log(2/\delta)}{n}}
\end{align*}
\end{proof}

\noindent\textbf{Remark}. This theorem provides probabilistic upper and lower bounds on the difference between the loss of the optimal solution to the constrained problem, $\mathcal{L}(\hat{\theta})$, and the loss of the optimal solution to the unconstrained problem, $\mathcal{L}(\theta^*)$. The upper bound shows that $\mathcal{L}(\hat{\theta})$ is at most $\frac{\epsilon}{2}$ plus a term that depends on the Lipschitz constant $L$, the number of samples $n$, and the confidence parameter $\delta$. The lower bound shows that $\mathcal{L}(\hat{\theta})$ is at least a negative term that depends on the same factors. These bounds hold with probability at least $1 - \delta$ and provide a way to quantify the effect of the fairness constraint on the performance of the learned model.

\begin{figure*}[!t]
	\centering
    \includegraphics[width=.9\textwidth]{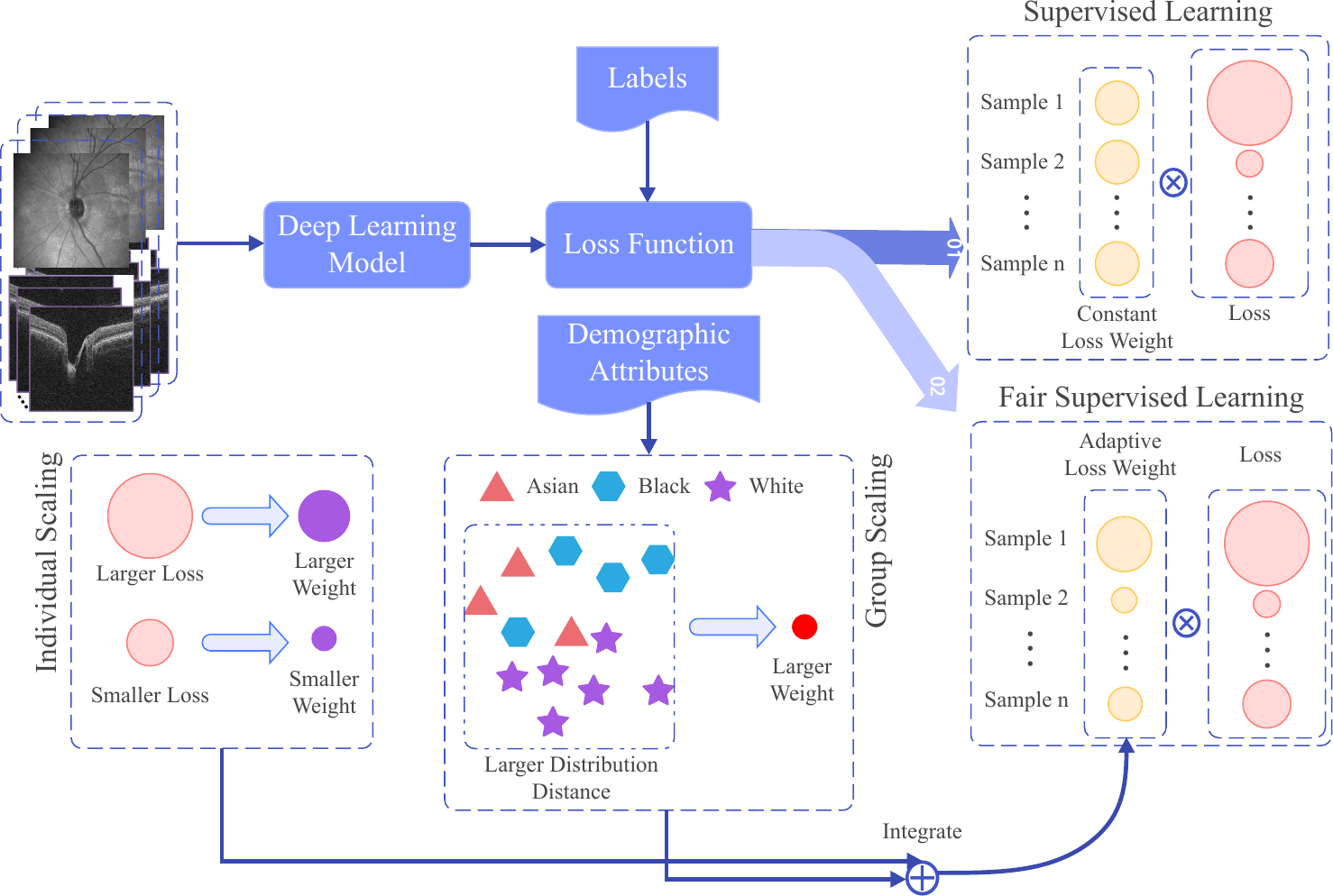} 
	\caption{\label{fig:schematic} Schematic view of the proposed FIS, in comparison to the standard supervised learning framework. Given inputs such as SLO fundus images or OCT B-scans, the proposed FIS incorporates individual scaling and group scaling to determine adaptive loss weights. In the visualization, a large circle in the loss (or loss weight) indicates a high value for the instance loss (or loss weight).}
\end{figure*}

\subsection{Fair Identity Scaling} 
The proposed FIS, as illustrated in Figure \ref{fig:schematic}, incorporates both individual scaling and group scaling to determine loss weights. Individual scaling aims to prioritizes the samples with large or small losses by upweighting the corresponding loss weights during training, while de-prioritizing the samples with large or small losses by downweighting the corresponding loss weights. This mechanism encourages the model to focus on the samples that it has not yet fully understood, promoting both learning flexibility and individual equity.
In a similar fashion, Group scaling aims to determine loss weights, based on distributions across demographic groups. It prioritizes the samples in a demographic group whose distribution deviates significantly from others by increasing their loss weights during training, while de-prioritizing samples in a demographic group with aligned distributions by reducing their loss weights. This mechanism encourages the model to focus on minority groups, thereby promoting group equity.

The aforementioned process can be mathematically formulated as 
\begin{align}
\mathcal{L}_{FIS}(\theta, \mathcal{B}^{t})=\frac{1}{|\mathcal{B}^{t}|}\sum_{(x_{i},y_{i},a_{i})\in \mathcal{B}^{t}}^{}
\left[((1-c)\cdot s^{I}_{i} + c \cdot s^{G}_{i}) \cdot \ell^{t}_{i} \right], 
\label{eqn:fair_scaling}
\end{align}
where $\mathcal{B}^{t}$ is a mini-batch at the training step t, $\ell^{t}_{i}$ is an instance-level loss, $s^{I}_{i}$ is the individual-scaled loss weight of the sample $(x_{i},y_{i},a_{i})$, and $s^{G}_{i}$ is the group-scaled loss weight of the sample $(x_{i},y_{i},a_{i})$. The fusion weight $c$ ranging between 0 and 1 is used to control group and individual scaling integration, where $c = 1$ means group scaling alone, and $c = 0$ indicates individual scaling alone.

Specifically, $s^{I}_{i}$ and $s^{G}_{i}$ are defined as
\begin{align}
s^{I}_{i} &= \frac{\exp( \ell^{t}_{i})}{\sum_{j=1}^{|\mathcal{B}^{t}|} \exp( \ell^{t}_{j})}, \label{eqn:ind_scaling}  \\
s^{G}_{i} &= \frac{\exp( \text{OT}(\{\ell^{t}\}|_{\mathcal{B}^{t}}, \{\ell^{t}\}|_{\mathcal{B}^{t}_{a_{i}}}) }{\sum_{j=1}^{|\mathcal{A}|} \exp( \text{OT}(\{\ell^{t}\}|_{\mathcal{B}^{t}}, \{\ell^{t}\}|_{\mathcal{B}^{t}_{a_{j}}}) )} \label{eqn:grp_scaling}
\end{align}
where $\text{OT}(\cdot, \cdot)$ is an optimal transport function, $\{\ell^{t}\}|_{\mathcal{B}^{t}}$ is a set of losses computed by the samples in $\mathcal{B}^{t}$, and $\mathcal{B}^{t}_{a_{j}}$ is a minibatch of the samples drawn from the set $\{(x,y,a)|(x,y,a) \in \mathcal{D}, a=a_{j}\}$.


\subsection{Theoretical Properties of FIS}

\begin{theorem}
\label{thm:bounds}
Let $\mathcal{L}(\theta, \mathcal{B}^{t})$ be the conventional loss and $\mathcal{L}_{FIS}(\theta, \mathcal{B}^{t})$ be the loss with the proposed FIS defined in Equation \ref{eqn:fair_scaling}. Assume that the individual loss $\ell^{t}_{i}$ is bounded, i.e., $0 \leq \ell^{t}_{i} \leq M$ for all $(x_{i},y_{i},a_{i}) \in \mathcal{B}^{t}$. Then, the following error bounds hold:

\begin{align*}
(1-c) \cdot \min_{i} s^{I}_{i} \cdot \mathcal{L}(\theta, \mathcal{B}^{t}) &\leq \mathcal{L}_{FIS}(\theta, \mathcal{B}^{t}) \\
&\leq (1-c) \cdot \max_{i} s^{I}_{i} \cdot \mathcal{L}(\theta, \mathcal{B}^{t}) + c \cdot M
\end{align*}
where $s^{I}_{i}$ is the individual-scaled loss weight defined in Equation \ref{eqn:ind_scaling}.
\end{theorem}

\begin{proof}
Starting from the definition of $\mathcal{L}_{FIS}(\theta, \mathcal{B}^{t})$ in Equation \ref{eqn:fair_scaling}:

\begin{align*}
& \mathcal{L}_{FIS}(\theta, \mathcal{B}^{t}) \\
&= \frac{1}{|\mathcal{B}^{t}|}\sum_{(x_{i},y_{i},a_{i})\in \mathcal{B}^{t}}^{} \left[((1-c)\cdot s^{I}_{i} + c \cdot s^{G}_{i}) \cdot \ell^{t}{i} \right] \ \\
&= (1-c) \cdot \frac{1}{|\mathcal{B}^{t}|}\sum_{(x_{i},y_{i},a_{i})\in \mathcal{B}^{t}}^{} s^{I}_{i} \cdot \ell^{t}{i} \\ 
&\hspace{5ex}+ c \cdot \frac{1}{|\mathcal{B}^{t}|}\sum_{(x_{i},y_{i},a_{i})\in \mathcal{B}^{t}}^{} s^{G}_{i} \cdot \ell^{t}{i}
\end{align*}

For the lower bound, we have:

\begin{align*}
\mathcal{L}_{FIS}(\theta, \mathcal{B}^{t}) &\geq (1-c) \cdot \frac{1}{|\mathcal{B}^{t}|}\sum_{(x_{i},y_{i},a_{i})\in \mathcal{B}^{t}}^{} s^{I}_{i} \cdot \ell^{t}{i} \ \\
&\geq (1-c) \cdot \min_{i} s^{I}_{i} \cdot \frac{1}{|\mathcal{B}^{t}|}\sum_{(x_{i},y_{i},a_{i})\in \mathcal{B}^{t}}^{} \ell^{t}_{i} \ \\
&= (1-c) \cdot \min_{i} s^{I}_{i} \cdot \mathcal{L}(\theta, \mathcal{B}^{t})
\end{align*}

For the upper bound, we have:

\begin{align*}
\mathcal{L}_{FIS}(\theta, \mathcal{B}^{t}) &\leq (1-c) \cdot \frac{1}{|\mathcal{B}^{t}|}\sum_{(x_{i},y_{i},a_{i})\in \mathcal{B}^{t}}^{} s^{I}_{i} \cdot \ell^{t}{i} \\ &\hspace{5ex}+ c \cdot \frac{1}{|\mathcal{B}^{t}|}\sum_{(x_{i},y_{i},a_{i})\in \mathcal{B}^{t}}^{} s^{G}_{i} \cdot \ell^{t}{i} \ \\
&\leq (1-c) \cdot \max_{i} s^{I}_{i} \cdot \frac{1}{|\mathcal{B}^{t}|}\sum_{(x_{i},y_{i},a_{i})\in \mathcal{B}^{t}}^{} \ell^{t}{i} \\ &\hspace{5ex}+ c \cdot \frac{1}{|\mathcal{B}^{t}|}\sum_{(x_{i},y_{i},a_{i})\in \mathcal{B}^{t}}^{} s^{G}_{i} \cdot M \ \\
&= (1-c) \cdot \max_{i} s^{I}_{i} \cdot \mathcal{L}(\theta, \mathcal{B}^{t}) \\ &\hspace{5ex}+ c \cdot M \cdot \frac{1}{|\mathcal{B}^{t}|}\sum_{(x_{i},y_{i},a_{i})\in \mathcal{B}^{t}}^{} s^{G}_{i} \ \\
&= (1-c) \cdot \max_{i} s^{I}_{i} \cdot \mathcal{L}(\theta, \mathcal{B}^{t}) + c \cdot M
\end{align*}
where the last equality holds because $\sum_{(x_{i},y_{i},a_{i})\in \mathcal{B}^{t}}^{} s^{G}_{i} \le |\mathcal{B}^{t}|$ by the definition of $s^{G}_{i}$ in Equation \ref{eqn:grp_scaling}.

Therefore, we have shown that:

\begin{align*}
(1-c) \cdot \min_{i} s^{I}_{i} \cdot \mathcal{L}(\theta, \mathcal{B}^{t}) &\leq \mathcal{L}_{FIS}(\theta, \mathcal{B}^{t}) \\ &\leq (1-c) \cdot \max_{i} s^{I}_{i} \cdot \mathcal{L}(\theta, \mathcal{B}^{t}) + c \cdot M
\end{align*}
which completes the proof.
\end{proof}

The theorem provides upper and lower bounds for the FIS loss $\mathcal{L}_{FIS}(\theta, \mathcal{B}^{t})$ in terms of the conventional loss $\mathcal{L}(\theta, \mathcal{B}^{t})$. The lower bound shows that $\mathcal{L}_{FIS}(\theta, \mathcal{B}^{t})$ is at least a scaled version of the conventional loss, where the scaling factor depends on the minimum individual-scaled loss weight $\min_{i} s^{I}_{i}$ and the fusion weight $(1-c)$. This suggests that the FIS method preserves the overall minimization objective of the conventional loss to a certain extent.

The upper bound, on the other hand, shows that $\mathcal{L}_{FIS}(\theta, \mathcal{B}^{t})$ is bounded by a scaled version of the conventional loss plus a constant term $c \cdot M$. The scaling factor in the upper bound depends on the maximum individual-scaled loss weight $\max_{i} s^{I}_{i}$ and the fusion weight $(1-c)$. The constant term $c \cdot M$ arises from the group-scaled loss weights $s^{G}_{i}$ and the assumption that the individual losses are bounded by $M$.

These bounds provide insights into the behavior of the proposed FIS compared to the conventional loss. The lower bound ensures that the FIS method does not completely deviate from the conventional loss, while the upper bound suggests that the FIS method can introduce additional terms to address fairness considerations. The tightness of these bounds depends on the specific values of the individual-scaled loss weights $s^{I}_{i}$, the group-scaled loss weights $s^{G}_{i}$, and the fusion weight $c$.

\begin{corollary}
\label{thm:FIS_outperform_base}
Theorem \ref{thm:bounds} suggests that  $\mathcal{L}_{FIS}(\theta, \mathcal{B}^{t})$ is lower than the conventional loss $\mathcal{L}(\theta, \mathcal{B}^{t})$ if either of the following conditions is satisfied:

\begin{enumerate}
    \item When $c = 0$ (individual scaling only): $\min_{i} s^{I}_{i} < 1$
    \item When $0 < c < 1$ (combination of individual and group scaling): $\min_{i} s^{I}_{i} < \frac{1}{1-c}$
\end{enumerate}
where $s^{I}_{i}$ is the individual-scaled loss weight defined in Equation \ref{eqn:ind_scaling}, and $c$ is the fusion weight controlling the balance between individual and group scaling.
\end{corollary}

\begin{proof}
From the lower bound in Theorem \ref{thm:bounds}, we have:

\begin{align*}
\mathcal{L}_{FIS}(\theta, \mathcal{B}^{t}) &\geq (1-c) \cdot \min_{i} s^{I}_{i} \cdot \mathcal{L}(\theta, \mathcal{B}^{t})
\end{align*}

For $\mathcal{L}_{FIS}(\theta, \mathcal{B}^{t})$ to be lower than $\mathcal{L}(\theta, \mathcal{B}^{t})$, we need:

\begin{align*}
(1-c) \cdot \min_{i} s^{I}_{i} \cdot \mathcal{L}(\theta, \mathcal{B}^{t}) &< \mathcal{L}(\theta, \mathcal{B}^{t})
\end{align*}

Simplifying this condition, we get:

\begin{align*}
(1-c) \cdot \min_{i} s^{I}_{i} &< 1
\end{align*}

Now, let's consider the two cases mentioned in the corollary:

\textbf{Case 1:} When $c = 0$ (individual scaling only):

In this case, the condition becomes:

\begin{align*}
\min_{i} s^{I}_{i} &< 1
\end{align*}

If the minimum individual-scaled loss weight $\min_{i} s^{I}_{i}$ is less than 1, then $\mathcal{L}_{FIS}(\theta, \mathcal{B}^{t})$ is guaranteed to be lower than $\mathcal{L}(\theta, \mathcal{B}^{t})$.

\textbf{Case 2:} When $0 < c < 1$ (combination of individual and group scaling):

In this case, the condition becomes:

\begin{align*}
(1-c) \cdot \min_{i} s^{I}_{i} &< 1 \\
\min_{i} s^{I}_{i} &< \frac{1}{1-c}
\end{align*}

If the minimum individual-scaled loss weight $\min_{i} s^{I}_{i}$ is less than $\frac{1}{1-c}$, then $\mathcal{L}_{FIS}(\theta, \mathcal{B}^{t})$ is guaranteed to be lower than $\mathcal{L}(\theta, \mathcal{B}^{t})$.

Therefore, if either of the conditions mentioned in the corollary is satisfied, the Fair Identity Scaling loss $\mathcal{L}_{FIS}(\theta, \mathcal{B}^{t})$ is guaranteed to be lower than the conventional loss $\mathcal{L}(\theta, \mathcal{B}^{t})$.
\end{proof}

Corollary \ref{thm:FIS_outperform_base} provides sufficient conditions under which the Fair Identity Scaling loss is lower than the conventional loss. These conditions depend on the minimum individual-scaled loss weight $\min_{i} s^{I}_{i}$ and the fusion weight $c$.

When $c = 0$ (individual scaling only), if there exists at least one sample in the batch whose individual-scaled loss weight is less than 1, then the FIS loss is guaranteed to be lower than the conventional loss.

When $0 < c < 1$ (combination of individual and group scaling), the condition becomes more relaxed as $c$ increases. As more importance is given to group scaling, the threshold for $\min_{i} s^{I}_{i}$ becomes larger, making it easier for the FIS loss to be lower than the conventional loss.

It is important to note that these conditions are sufficient but not necessary for $\mathcal{L}_{FIS}(\theta, \mathcal{B}^{t})$ to be lower than $\mathcal{L}(\theta, \mathcal{B}^{t})$. In practice, the actual behavior of the FIS loss will depend on the specific values of the individual-scaled loss weights $s^{I}_{i}$, the group-scaled loss weights $s^{G}_{i}$, and the fusion weight $c$.

	\section{Experimental Setup}




\noindent\textbf{Dataset Splits:} For each of the AMD, DR, and Glaucoma datasets, we have 10,000 samples from 10,000 subjects. We use 6,000, 1,000, and 3,000 samples for training, validation, and testing respectively and report the mean and standard deviation across 5 runs.

\noindent\textbf{Models:} We conduct a comprehensive evaluation of overall model performance and fairness across 5 different architectures encompassing both CNNs and transformers. Specifically, we leverage the ResNet50~\cite{he2016deep}, DenseNet121~\cite{huang2017densely}, and EfficientNet~\cite{tan2019efficientnet} CNNs and the ViT-B~\cite{dosovitskiy2020image} and Swin-B~\cite{liu2021swin} transformer architectures. For training via the 3D OCT B-Scans, we utilize a 3D ResNet model following \cite{Yang_JBHI_2021}.

\noindent\textbf{Training Scheme:}  The CNN models, \ie, ResNet50, DenseNet121, and EfficientNet are trained with a learning rate of 1e-4 for 10 epochs with a batch size of 6. Swin-B is trained with a learning rate of 1e-4 for 50 epochs using a batch size of 64. Lastly, for ViT-B, we follow the MAE~\cite{he2022masked} settings and train for 50 epochs with a layer decay = 0.55, weight decay = 0.01, drop path rate = 0.1, batch size = 64, and base learning rate = 5e-4. For training via the 3D OCT B-scans, we utilize a learning rate of 5e-5 for 50 epochs with a batch size of 2. AdamW~\cite{Loshchilov_ICLR_2019} is used as the optimization method for all models.

For the proposed FIS, we set the fusion weight $c$ to 0.5. For the adversarial fair loss, we set $\lambda=0.2$ to ensure all loss terms align on a similar scale, and we maintain the default settings for other experimental aspects. For the pretraining of FSCL, we utilize the default hyper-parameters and experimental setups provided in \cite{wang2022fairness}.

\noindent\textbf{Metrics:} To facilitate the model fairness assessment, we report overall AUC, equity-scaled AUC~\cite{Luo_TMI_2024}, as well as the group-wise AUCs across the protected attributes of race, gender, and ethnicity. In addition, we use traditional fairness metrics of DPD and DEOdds to assess model fairness. Furthermore, we propose to use our performance-scaled disparity (Mean and Max PSDs) scores to evaluate model fairness in the context of overall model performance. Specifically, they are defined as
\begin{align*}
\small
    \text{Mean PSD} &= \sqrt{\frac{\sum_{a\in \mathcal{A}}\left(\text{AUC}_{a}-\text{AUC}_{mean}\right)^2}{N}}/\text{Overall AUC} \\ 
    \text{Max PSD} &= (\text{AUC}_{a_{max}}-\text{AUC}_{a_{min}})/\text{Overall AUC}
\end{align*}
where $\text{AUC}_{mean}$ is the average AUC over all the groups, and $a_{max}$ ($a_{min}$) is the group achieving the largest (smallest) AUC across the groups.

\noindent\textbf{Fairness Methods:} We choose three SOTA fairness methods including fair adversarial training (Adv)~\cite{beutel2017data}, GroupDRO \cite{sagawa2019distributionally}, and fair contrastive loss~\cite{wang2022fairness} (FSCL) for comparison with our proposed FIS method. Since the FSCL model is based on image augmentations, we only use it with 2D fundus images since effective image augmentation strategies for 3D imaging data are largely unclear in the literature. Similarly, GroupDRO is only applied in the experiments on SLO fundus images. We follow the official code repositories, experimental protocols and hyper-parameter choices for these fairness methods.

\section{Results}
Here, we present a thorough evaluation of model fairness in Section~\ref{sec:fairness_benchmark}, which uncovers significant biases across multiple protected attributes. To address these fairness gaps, Section~\ref{sec:fis_results} provides results of the proposed FIS method, illustrating its effectiveness in improving fairness.

\subsection{Fairness Benchmark -- Harvard-FairVision}
\label{sec:fairness_benchmark}
In this section, we present a comprehensive evaluation of model performance and fairness across 5 different architectures trained on SLO fundus images from our Harvard-FairVision dataset. Specifically, Tables~\ref{tbl:slo_race_archs},~\ref{tbl:slo_gender_archs}, and~\ref{tbl:slo_ethnicity_archs} report results across the protected attributes of race, gender, and ethnicity respectively. In addition to the overall performance measured via Overall AUC, we also report the fairness metrics (Mean PSD, Max PSD, DEOdds, and ES-AUC) as well as the group-wise AUCs. In terms of Overall AUC, we note that EfficientNet yields the best results out of the CNN models on AMD and DR detection with AUCs of 79.50\% and 80.19\% respectively, whereas DenseNet121 obtains best results on Glaucoma detection with an AUC of 78.76\%. However, the transformer models ViT-B and Swin-B consistently outperform these CNN models, with ViT-B achieving the best results on DR and Glaucoma detection with AUCs of 85.22\% and 79.73\% respectively, whereas Swin-B outperforming on AMD detection with an AUC of 82.58\%. Next, we study the fairness of these 5 models across the protected attributes of race, gender, and ethnicity respectively.

\noindent\textbf{Race Fairness:} Across race (Table~\ref{tbl:slo_race_archs}), ViT-B obtains the best performance-fairness trade-off on AMD and DR detection with ES-AUCs of 74.56\% and 78.66\% respectively whereas Swin-B exhibits best results on Glaucoma detection with an ES-AUC of 76.05\%. In terms of the individual subgroups, we note significant disparities in model performance. For instance, on AMD detection, all 5 models consistently obtain the best performance on the White subgroup and the worst performance on the Black subgroup. On DR detection, the White subgroup is again favored, with the least favored subgroup being either Asian or Black depending on the model architecture. On Glaucoma detection, all 5 models consistently overperform on the Asian subgroup and consistently underperform on the Black subgroup. Hence, overall, we observe that the White subgroup is favored on AMD and DR detection whereas the Asian subgroup is favored on Glaucoma detection.

\noindent\textbf{Gender Fairness:} Across gender (Table~\ref{tbl:slo_gender_archs}), ViT-B obtains the best performance-fairness trade-off on DR detection with an ES-AUC of 84.60\% whereas Swin-B exhibits best results on AMD and Glaucoma detection with ES-AUCs of 81.09\% and 78.19\% respectively. In terms of the individual Female and Male subgroups, we again observe disparities in model performance, as reflected by the group-wise AUCs. On AMD detection, all 5 models consistently yield the best performance on Female patients and worst on Male patients. On DR detection, the CNN models favor Female patients whereas the transformer models favor Male patients. On the other hand, on Glaucoma detection, all 5 models consistently overperform on Male patients compared to Female patients. Overall, we note that the Female subgroup is favored on AMD detection whereas the Male subgroup is favored on Glaucoma detection, with CNNs and transformers favoring different subgroups on DR detection.

\noindent\textbf{Ethnicity Fairness:} Across ethnicity (Table~\ref{tbl:slo_ethnicity_archs}), ViT-B achieves the best performance-fairness trade-off on AMD and DR detection with ES-AUCs of 80.13\% and 82.18\% respectively whereas DenseNet121 exhibits best results on Glaucoma detection with an ES-AUC of 75.60\%. In terms of the individual subgroups, we again notice disparities in model performance between the non-Hispanic and Hispanic subgroups. On AMD detection, all models except ViT-B yield better results on non-Hispanic patients. On the other hand, on DR detection, all models except ResNet50 favor the Hispanic subgroup. Lastly, on Glaucoma detection, we note significant disparities between the non-Hispanic and Hispanic subgroups, with all 5 models consistently overperforming on the former subgroup. Overall, we observe that the non-Hispanic subgroup is favored on AMD and Glaucoma detection whereas the Hispanic subgroup is favored on DR detection.

\noindent\textbf{Summary:} To summarize, our comprehensive results on 5 different architectures and 3 eye diseases reveal significant disparities in performance across the protected attributes of race, gender, and ethnicity, with ViT-B achieving the best performance-fairness trade-off. Hence, we primarily focus on the ViT-B architecture for the subsequent analyses and propose our Fair Identity Scaling to alleviate these biases.

\subsection{Fair Identity Scaling (FIS)}
\label{sec:fis_results}
\begin{table*}[!t]
	\centering
	\caption{
	    Performance and fairness across \textbf{race} of different architectures trained on SLO fundus images from the Harvard-FairVision dataset. Mean and standard deviation are reported across five runs.
	}
	\adjustbox{width=1\textwidth}{
	\begin{tabular}{C{10ex} L{18ex} C{12ex} C{12ex} C{12ex} C{12ex} C{12ex} C{12ex} C{12ex} C{12ex}}
		\toprule
          \multirow{2}{*}{\textbf{Disease}} & \multirow{2}{*}{\textbf{Method}} & \textbf{Overall} & \multirow{2}{*}{\textbf{ES-AUC$\uparrow$}} & \textbf{Asian} & \textbf{Black} & \textbf{White} & \textbf{Mean} & \textbf{Max} &  \textbf{\multirow{2}{*}{\textbf{DEOdds$\downarrow$}}} \\
	 & & \textbf{AUC$\uparrow$} & & \textbf{AUC$\uparrow$} & \textbf{AUC$\uparrow$} & \textbf{AUC$\uparrow$} & \textbf{PSD$\downarrow$} & \textbf{PSD$\downarrow$} & \\
		\cmidrule(lr){1-1} \cmidrule(lr){2-2} \cmidrule(lr){3-4}  \cmidrule(lr){5-7} \cmidrule(lr){8-10}

        \multirow{5}{*}{\textbf{AMD}} & ResNet50 & 76.64$\pm$0.42 & 67.41$\pm$2.02 & 71.78$\pm$4.39 &	68.85$\pm$5.40 &	75.89$\pm$0.31  & 5.38$\pm$2.25	& 12.35$\pm$4.99  & \textbf{39.26}$\pm$3.85 \\
        & DenseNet121 & 79.14$\pm$0.54 & 70.54$\pm$3.55 & 74.90$\pm$2.35	& 71.85$\pm$4.40 &	78.28$\pm$0.41  & 3.63$\pm$1.77	& 8.68$\pm$4.41	&	47.70$\pm$16.78  \\
        & EfficientNet & 79.50$\pm$0.44 & 71.06$\pm$0.62 & 77.22$\pm$1.25	& 70.68$\pm$1.67 &	78.71$\pm$0.48  & 4.43$\pm$1.26	& 10.12$\pm$2.38	&	43.74$\pm$5.06    \\
        & ViT-B & 82.32$\pm$0.17 & \textbf{74.56}$\pm$0.87 & 79.01$\pm$0.94 &	\textbf{76.13}$\pm$1.04 &	81.40$\pm$0.15     & \textbf{2.67}$\pm$0.58 & \textbf{6.40}$\pm$1.36    & 42.38$\pm$2.83   \\
        & Swin-B & \textbf{82.58}$\pm$0.19 & 70.63$\pm$1.55 & \textbf{79.43}$\pm$0.83	& 69.17$\pm$3.09	& \textbf{82.18}$\pm$0.22     & 6.81$\pm$1.86	& 15.76$\pm$3.94		& 47.96$\pm$8.21       \\ \midrule

        \multirow{4}{*}{\textbf{DR}} & ResNet50 & 76.36$\pm$1.05 & 60.71$\pm$1.40 & 54.76$\pm$4.32 &	74.32$\pm$2.54 &	77.98$\pm$1.39  & 13.44$\pm$3.25 &	30.44$\pm$5.97	&	20.58$\pm$9.79  \\
        & DenseNet121 & 78.18$\pm$1.07 & 61.59$\pm$2.82 & 62.15$\pm$3.00	& 69.82$\pm$2.49 &	80.95$\pm$1.86   & 9.97$\pm$1.94	& 24.03$\pm$4.71	&	27.02$\pm$8.48  \\
        & EfficientNet & 80.19$\pm$0.57 & 69.57$\pm$2.61 & 73.49$\pm$3.82	& 73.20$\pm$1.72	& 81.87$\pm$0.26  & 5.33$\pm$1.37	& 12.62$\pm$3.48 &	\textbf{19.33}$\pm$8.51  \\
        & ViT-B & \textbf{85.22}$\pm$0.21 & \textbf{78.66}$\pm$1.76 & \textbf{83.41}$\pm$1.38 &	\textbf{79.69}$\pm$0.89 &	\textbf{86.27}$\pm$0.40     & \textbf{3.22}$\pm$0.56 &	\textbf{7.72}$\pm$1.40	& 	21.51$\pm$7.14       \\
        & Swin-B & 81.87$\pm$0.44 & 66.83$\pm$2.26 & 65.74$\pm$4.17 &	77.12$\pm$2.11	& 83.61$\pm$0.81     & 9.17$\pm$2.46 &	21.82$\pm$5.36 	& 20.63$\pm$4.69       \\ \midrule

        \multirow{4}{*}{\textbf{Glaucoma}} & ResNet50 & 77.07$\pm$0.46 & 72.60$\pm$2.44 & 80.32$\pm$2.20	& 74.32$\pm$2.06	& 77.04$\pm$0.44  & 3.25$\pm$1.88	& 7.89$\pm$4.60	&	8.38$\pm$4.41  \\
        & DenseNet121 & 78.76$\pm$0.74 & 73.98$\pm$1.13 & 81.10$\pm$2.26	& 74.98$\pm$1.45 &	79.13$\pm$0.72  & 3.33$\pm$0.98	& 7.79$\pm$2.46	&	9.45$\pm$1.16  \\
        & EfficientNet & 78.20$\pm$0.43 & 72.50$\pm$0.50 & 81.57$\pm$0.61	& 73.97$\pm$0.36	& 78.44$\pm$0.58  & 4.03$\pm$0.14	& 9.72$\pm$0.47	&	11.10$\pm$2.23  \\
        & ViT-B & \textbf{79.73}$\pm$0.18 & 74.25$\pm$1.08 & \textbf{83.78}$\pm$0.85 &	76.46$\pm$0.70	& \textbf{79.69}$\pm$0.14     & 3.76$\pm$0.72 &	9.18$\pm$1.73	& 9.45$\pm$2.28       \\
        & Swin-B & 79.07$\pm$0.49 & \textbf{76.05}$\pm$1.37 & 81.23$\pm$0.75	& \textbf{77.54}$\pm$1.29	& 79.02$\pm$0.77     & \textbf{2.03}$\pm$0.87 &	\textbf{4.76}$\pm$2.11	 & \textbf{5.49}$\pm$1.88      \\
        
		\bottomrule	
	\end{tabular}}
\label{tbl:slo_race_archs}
\end{table*}

\begin{table*}[!t]
	\centering
	\caption{
	    Performance and fairness across \textbf{gender} of different architectures trained on SLO fundus images from the Harvard-FairVision dataset. Mean and standard deviation are reported across five runs.
	}
	\adjustbox{width=1\textwidth}{
	\begin{tabular}{C{10ex} L{18ex} C{12ex} C{12ex} C{12ex} C{12ex} C{12ex} C{12ex} C{12ex} C{12ex}}
		\toprule
          \multirow{2}{*}{\textbf{Disease}} & \multirow{2}{*}{\textbf{Method}} & \textbf{Overall} & \multirow{2}{*}{\textbf{ES-AUC$\uparrow$}} & \textbf{Female} & \textbf{Male} & \textbf{Mean} & \textbf{Max} &  \textbf{\multirow{2}{*}{\textbf{DEOdds$\downarrow$}}} \\
	 & & \textbf{AUC$\uparrow$} & & \textbf{AUC$\uparrow$} & \textbf{AUC$\uparrow$} & \textbf{PSD$\downarrow$} & \textbf{PSD$\downarrow$} & \\
		\cmidrule(lr){1-1} \cmidrule(lr){2-2} \cmidrule(lr){3-4}  \cmidrule(lr){5-6} \cmidrule(lr){7-9}

        \multirow{4}{*}{\textbf{AMD}} & ResNet50 & 76.64$\pm$0.42 & 75.56$\pm$0.65 & 77.14$\pm$0.86 &	75.70$\pm$0.52   & \textbf{0.94}$\pm$0.84 &	\textbf{1.88}$\pm$1.68	& \textbf{7.49}$\pm$1.81   \\
        & DenseNet121 & 79.14$\pm$0.54 & 76.51$\pm$0.45 & 80.41$\pm$0.66	& 76.95$\pm$0.46  & 2.18$\pm$0.34	& 4.36$\pm$0.68	&	10.31$\pm$1.57   \\
        & EfficientNet & 79.50$\pm$0.44 & 77.28$\pm$0.37 & 80.54$\pm$0.59	& 77.68$\pm$0.36  & 1.80$\pm$0.33	& 3.60$\pm$0.66	&	9.88$\pm$2.41   \\
        & ViT-B & 82.32$\pm$0.17 & 80.34$\pm$0.26 & 83.23$\pm$0.25 &	80.77$\pm$0.22   & 1.50$\pm$0.22 &	3.00$\pm$0.44		& 10.95$\pm$3.04        \\
        & Swin-B & \textbf{82.58}$\pm$0.19 & \textbf{81.09}$\pm$0.64 & \textbf{83.30}$\pm$0.24 &	\textbf{81.45}$\pm$0.54   & 1.11$\pm$0.41 &	2.23$\pm$0.82		& 16.63$\pm$9.80       \\ \midrule

        \multirow{4}{*}{\textbf{DR}} & ResNet50 & 76.36$\pm$1.05 & 74.67$\pm$1.65 & 76.77$\pm$1.23	& 75.92$\pm$2.01 &  1.50$\pm$0.92	& 3.00$\pm$1.84	&	\textbf{2.49}$\pm$2.23   \\
        & DenseNet121 & 78.18$\pm$1.07 & 76.09$\pm$1.48 & 78.89$\pm$2.13	& 77.26$\pm$1.42  & 1.77$\pm$0.76	& 3.54$\pm$1.52	&	2.59$\pm$0.95  \\
        & EfficientNet & 80.19$\pm$0.57 & 78.73$\pm$0.97 & 81.04$\pm$1.16	& 79.37$\pm$0.68  & 1.16$\pm$0.73	& 2.32$\pm$1.46	&	3.01$\pm$1.73  \\
        & ViT-B & \textbf{85.22}$\pm$0.21 & \textbf{84.60}$\pm$0.64 & \textbf{84.84}$\pm$0.54 &	\textbf{85.43}$\pm$0.44   & \textbf{0.43}$\pm$0.42 &	\textbf{0.86}$\pm$0.83  &	5.48$\pm$4.40       \\
        & Swin-B & 81.87$\pm$0.44 & 81.25$\pm$0.58 & 81.47$\pm$0.68 &	82.01$\pm$0.45   & 0.46$\pm$0.30	& 0.93$\pm$0.60  &	6.69$\pm$3.53        \\ \midrule

        \multirow{4}{*}{\textbf{Glaucoma}} & ResNet50 & 77.07$\pm$0.46 & 76.03$\pm$0.63 & 76.49$\pm$0.48	& 77.79$\pm$0.86  & 0.89$\pm$0.53	& 1.79$\pm$1.06	&	3.33$\pm$2.41   \\
        & DenseNet121 & 78.76$\pm$0.74 & 77.75$\pm$0.54 & 78.19$\pm$0.57	& 79.49$\pm$0.98  & 0.82$\pm$0.39	& 1.65$\pm$0.78	&	3.49$\pm$3.54   \\
        & EfficientNet & 78.20$\pm$0.43 & 76.78$\pm$0.47 & 77.38$\pm$0.37	& 79.23$\pm$0.67  & 1.18$\pm$0.41	& 2.36$\pm$0.81	&	3.22$\pm$0.89   \\
        & ViT-B & \textbf{79.73}$\pm$0.18 & 77.80$\pm$0.29 & \textbf{78.63}$\pm$0.18 &	\textbf{81.11}$\pm$0.38   & 1.55$\pm$0.29	& 3.10$\pm$0.58  &	\textbf{2.66}$\pm$0.78        \\
        & Swin-B & 79.07$\pm$0.49 & \textbf{78.19}$\pm$0.80 & 78.53$\pm$0.66 &	79.66$\pm$0.33   & \textbf{0.72}$\pm$0.32 &	\textbf{1.44}$\pm$0.64		& 6.25$\pm$1.75        \\
        
		\bottomrule	
	\end{tabular}}
\label{tbl:slo_gender_archs}
\end{table*}

\begin{table*}[!t]
	\centering
	\caption{
	    Performance and fairness across \textbf{ethnicity} of different architectures trained on SLO fundus images from the Harvard-FairVision dataset. Mean and standard deviation are reported across five runs.
	}
	\adjustbox{width=1\textwidth}{
	\begin{tabular}{C{10ex} L{18ex} C{12ex} C{12ex} C{12ex} C{12ex} C{12ex} C{12ex} C{12ex} C{12ex}}
		\toprule
          \multirow{2}{*}{\textbf{Disease}} & \multirow{2}{*}{\textbf{Method}} & \textbf{Overall} & \multirow{2}{*}{\textbf{ES-AUC$\uparrow$}} & \textbf{Non-Hisp} & \textbf{Hispanic} & \textbf{Mean} & \textbf{Max} &  \textbf{\multirow{2}{*}{\textbf{DEOdds$\downarrow$}}} \\
	 & & \textbf{AUC$\uparrow$} & & \textbf{AUC$\uparrow$} & \textbf{AUC$\uparrow$} & \textbf{PSD$\downarrow$} & \textbf{PSD$\downarrow$} & \\
		\cmidrule(lr){1-1} \cmidrule(lr){2-2} \cmidrule(lr){3-4}  \cmidrule(lr){5-6} \cmidrule(lr){7-9}

        \multirow{4}{*}{\textbf{AMD}} & ResNet50 & 76.64$\pm$0.42 & 74.54$\pm$1.84 & 76.63$\pm$0.41	& 76.34$\pm$3.64  & 1.85$\pm$1.39	& 3.70$\pm$2.77	&	\textbf{18.89}$\pm$4.43   \\
        & DenseNet121 & 79.14$\pm$0.54 & 77.33$\pm$1.27 & 78.64$\pm$1.26	& 78.23$\pm$2.97  & 1.49$\pm$1.35	& 2.98$\pm$2.71	&	23.95$\pm$5.10   \\
        & EfficientNet & 79.50$\pm$0.44 & 77.87$\pm$1.36 & 79.55$\pm$0.40	& 77.61$\pm$1.90  & \textbf{1.33}$\pm$0.88	& \textbf{2.65}$\pm$1.76	&	23.18$\pm$5.64   \\
        & ViT-B & 82.32$\pm$0.17 & \textbf{80.13}$\pm$0.95 & 82.24$\pm$0.13	& \textbf{84.99}$\pm$1.48   & 1.67$\pm$0.84 &	3.33$\pm$1.68 	& 26.52$\pm$6.13        \\
        & Swin-B & \textbf{82.58}$\pm$0.19 & 77.74$\pm$1.18 & \textbf{82.73}$\pm$0.19 &	76.49$\pm$1.56   & 3.78$\pm$0.91	& 7.57$\pm$1.83 	& 21.36$\pm$6.44        \\ \midrule

        \multirow{4}{*}{\textbf{DR}} & ResNet50 & 76.36$\pm$1.05 & 74.08$\pm$1.53 & 76.38$\pm$1.05	& 74.42$\pm$3.22  & 2.02$\pm$0.96	& 4.04$\pm$1.92	&	14.26$\pm$9.93   \\
        & DenseNet121 & 78.18$\pm$1.07 & 75.72$\pm$1.99 & 77.95$\pm$1.01	& 80.10$\pm$3.76  & 2.10$\pm$1.32	& 4.21$\pm$2.63	&	13.33$\pm$5.41  \\
        & EfficientNet & 80.19$\pm$0.57 & 76.82$\pm$2.46 & 79.89$\pm$0.57	& 84.08$\pm$3.86  & 2.75$\pm$2.18	& 5.49$\pm$4.35	&	11.94$\pm$7.62  \\
        & ViT-B & \textbf{85.22}$\pm$0.21 & \textbf{82.18}$\pm$1.18 & \textbf{84.95}$\pm$0.24	& \textbf{88.65}$\pm$1.24   & 2.17$\pm$0.80 &	4.35$\pm$1.59		& 8.65$\pm$3.31       \\
        & Swin-B & 81.87$\pm$0.44 & 79.84$\pm$1.79 & 81.59$\pm$0.53 &	83.75$\pm$2.39   & \textbf{1.55}$\pm$1.34 &	\textbf{3.11}$\pm$2.68 	& \textbf{7.23}$\pm$2.44        \\ \midrule

        \multirow{4}{*}{\textbf{Glaucoma}} & ResNet50 & 77.07$\pm$0.46 & 71.04$\pm$1.86 & 77.36$\pm$0.45	& 68.81$\pm$2.68  & 5.54$\pm$1.75	& 11.09$\pm$3.50	&	10.95$\pm$2.82   \\
        & DenseNet121 & 78.76$\pm$0.74 & \textbf{75.60}$\pm$1.07 & 78.91$\pm$0.78	& \textbf{74.72}$\pm$1.40  & \textbf{2.66}$\pm$1.02	& \textbf{5.32}$\pm$2.04	&	6.66$\pm$1.12   \\
        & EfficientNet & 78.20$\pm$0.43 & 73.88$\pm$1.60 & 78.39$\pm$0.39	& 72.52$\pm$2.11  & 3.76$\pm$1.23	& 7.52$\pm$2.46	&	7.48$\pm$5.54   \\
        & ViT-B & \textbf{79.73}$\pm$0.18 & 73.95$\pm$0.94 & \textbf{80.00}$\pm$0.15	& 72.18$\pm$1.28   & 4.91$\pm$0.79 &	9.81$\pm$1.58		& 10.23$\pm$1.62        \\
        & Swin-B & 79.07$\pm$0.49 & 75.58$\pm$1.72 & 79.24$\pm$0.51 &	74.58$\pm$2.30   & 2.94$\pm$1.52	& 5.89$\pm$3.05  &	\textbf{4.50}$\pm$3.59        \\
        
		\bottomrule	
	\end{tabular}}
\label{tbl:slo_ethnicity_archs}
\end{table*}

\begin{table*}[!t]
	\centering
	\caption{Comparison of the proposed FIS method against other fairness baselines trained on \textbf{SLO fundus images} and evaluated on the protected attribute of \textbf{race}.
	}
	\adjustbox{width=1\textwidth}{
	\begin{tabular}{C{10ex} L{18ex} C{12ex} C{12ex} C{12ex} C{12ex} C{12ex} C{12ex} C{12ex} C{12ex}}
		\toprule
          \multirow{2}{*}{\textbf{Disease}} & \multirow{2}{*}{\textbf{Method}} & \textbf{Overall} & \textbf{\multirow{2}{*}{\textbf{ES-AUC$\uparrow$}}} & \textbf{Asian} & \textbf{Black} & \textbf{White} & \textbf{Mean} & \textbf{Max} & \textbf{\multirow{2}{*}{\textbf{DEOdds$\downarrow$}}} \\
	 & & \textbf{AUC$\uparrow$} &  & \textbf{AUC$\uparrow$} & \textbf{AUC$\uparrow$} & \textbf{AUC$\uparrow$} & \textbf{PSD$\downarrow$} & \textbf{PSD$\downarrow$} & \\
		\cmidrule(lr){1-1} \cmidrule(lr){2-2} \cmidrule(lr){3-3} \cmidrule(lr){4-4} \cmidrule(lr){5-7}  \cmidrule(lr){8-9} \cmidrule(lr){10-10}

        \multirow{4}{*}{\textbf{AMD}} & ViT & 82.32$\pm$0.17 & 74.51$\pm$0.87 & 79.01$\pm$0.94 & 76.13$\pm$1.04 & 81.40$\pm$0.15 & 2.72$\pm$0.58 & 6.45$\pm$1.36 & 42.38$\pm$2.83   \\
        & ViT+Adv & 82.15$\pm$0.51 & 74.53$\pm$0.75 & 78.96$\pm$0.37 & 76.05$\pm$1.21 & 81.23$\pm$0.11 & 2.61$\pm$0.67 & 6.31$\pm$1.64 & 38.65$\pm$6.16 \\
        & ViT+GroupDRO & 81.65$\pm$0.10 & 73.13$\pm$0.61 & 78.86$\pm$0.45 & 73.67$\pm$1.16 & 80.71$\pm$0.12 & 3.67$\pm$0.69 & 8.64$\pm$1.38 & \textbf{30.83}$\pm$8.51 \\
        & ViT+FSCL & 82.22$\pm$0.36 & 74.62$\pm$0.61 & 79.42$\pm$0.54 & 75.82$\pm$1.33 & 81.23$\pm$0.35 & 2.73$\pm$0.79 & 6.58$\pm$1.22 & 35.62$\pm$5.74  \\
        & ViT+FIS & \textbf{82.65}$\pm$0.39 & \textbf{76.64}$\pm$2.28 & \textbf{80.19}$\pm$1.04 & \textbf{78.04}$\pm$2.36 & \textbf{81.81}$\pm$0.43 & \textbf{1.89}$\pm$1.19 & \textbf{4.57}$\pm$2.88 & 35.47$\pm$1.68 \\ \midrule

        \multirow{4}{*}{\textbf{DR}} & ViT & 85.22$\pm$0.21 & 78.62$\pm$1.76 & 83.41$\pm$1.38 & 79.69$\pm$0.89 & 86.27$\pm$0.40 & 3.20$\pm$0.56 & 7.69$\pm$1.40 & 21.51$\pm$7.14  \\
        & ViT+Adv & 84.87$\pm$0.31 & 78.22$\pm$0.46 & 81.96$\pm$0.41 & 80.23$\pm$0.72 & 85.83$\pm$0.31 & 2.78$\pm$0.32 & 6.61$\pm$0.98 & 16.23$\pm$2.46   \\
        & ViT+GroupDRO & 84.62$\pm$0.49 & 78.33$\pm$0.27 & 81.55$\pm$0.59 & 80.69$\pm$0.93 & 85.55$\pm$0.66 & \textbf{2.53}$\pm$0.77 & \textbf{5.80}$\pm$1.07 & 18.93$\pm$3.15   \\
        & ViT+FSCL & 85.29$\pm$0.62 & 78.69$\pm$0.39 & 82.29$\pm$0.44 & 80.75$\pm$1.07 & 86.18$\pm$0.72 & 2.66$\pm$0.82 & 6.32$\pm$1.54 & \textbf{13.57}$\pm$2.85   \\
        & ViT+FIS & \textbf{85.71}$\pm$0.41 & \textbf{78.96}$\pm$0.53 & \textbf{83.66}$\pm$1.22 & \textbf{80.89}$\pm$0.87 & \textbf{87.01}$\pm$0.47 & 2.63$\pm$0.86 & 6.55$\pm$1.12 & 19.85$\pm$5.61 \\ \midrule

        \multirow{4}{*}{\textbf{Glaucoma}} & ViT & 79.73$\pm$0.18 & 74.23$\pm$1.08 & 83.78$\pm$0.85 & 76.46$\pm$0.70 & 79.69$\pm$0.14 & 3.79$\pm$0.72 & 9.19$\pm$1.73 & 9.45$\pm$2.28  \\
        & ViT+Adv & 79.73$\pm$0.29 & 73.71$\pm$0.81 & 84.07$\pm$0.40 & 76.00$\pm$0.70 & 79.77$\pm$0.06 & 4.13$\pm$0.39 & 10.13$\pm$0.95 & \textbf{6.05}$\pm$2.17  \\
        & ViT+GroupDRO & 79.21$\pm$0.33 & 74.25$\pm$1.26 & 83.77$\pm$0.29 & 77.83$\pm$0.58 & 78.47$\pm$0.12 & 3.36$\pm$1.55 & 7.49$\pm$2.48 & 8.35$\pm$1.93  \\
        & ViT+FSCL & 80.33$\pm$0.28 & 74.03$\pm$0.85 & 84.15$\pm$0.64 & 78.26$\pm$0.78 & 77.71$\pm$0.39 & 3.62$\pm$1.78 & 8.01$\pm$1.87 & 9.56$\pm$5.28  \\
        & ViT+FIS & \textbf{80.65}$\pm$0.23 & \textbf{76.25}$\pm$0.38 & \textbf{83.83}$\pm$0.41 & \textbf{77.66}$\pm$0.34 & \textbf{80.53}$\pm$0.25 & \textbf{2.85}$\pm$0.16 & \textbf{7.11}$\pm$0.38 & 9.41$\pm$2.67 \\ 
        
		\bottomrule	
	\end{tabular}}
\label{tbl:slo_race_fis}
\end{table*}

\begin{table*}[!t]
	\centering
	\caption{
        Comparison of the proposed FIS method against other fairness baselines trained on \textbf{OCT B-Scans} and evaluated on the protected attribute of \textbf{race}.
	}
	\adjustbox{width=1\textwidth}{
	\begin{tabular}{C{10ex} L{18ex} C{12ex} C{12ex} C{12ex} C{12ex} C{12ex} C{12ex} C{12ex} C{12ex}}
		\toprule
          \multirow{2}{*}{\textbf{Disease}} & \multirow{2}{*}{\textbf{Method}} & \textbf{Overall} & \textbf{\multirow{2}{*}{\textbf{ES-AUC$\uparrow$}}} & \textbf{Asian} & \textbf{Black} & \textbf{White} & \textbf{Mean} & \textbf{Max} & \textbf{\multirow{2}{*}{\textbf{DEOdds$\downarrow$}}} \\
	 & & \textbf{AUC$\uparrow$} &  & \textbf{AUC$\uparrow$} & \textbf{AUC$\uparrow$} & \textbf{AUC$\uparrow$} & \textbf{PSD$\downarrow$} & \textbf{PSD$\downarrow$} & \\
		\cmidrule(lr){1-1} \cmidrule(lr){2-2} \cmidrule(lr){3-3} \cmidrule(lr){4-4} \cmidrule(lr){5-7}  \cmidrule(lr){8-9} \cmidrule(lr){10-10}

        \multirow{3}{*}{\textbf{AMD}} & 3D ResNet & 87.63$\pm$0.29 & 83.93$\pm$1.93 & 88.35$\pm$1.44 & 84.30$\pm$2.26 & 87.28$\pm$0.24 & 1.95$\pm$0.79 & 4.64$\pm$1.68 & 28.51$\pm$3.73 \\
        & 3D ResNet+Adv & 86.88$\pm$0.38 & 80.17$\pm$1.05 & 86.97$\pm$0.69 & 79.08$\pm$1.54 & 86.40$\pm$0.38 & 4.14$\pm$1.07 & 9.06$\pm$2.42 & 33.21$\pm$6.12 \\
        & 3D ResNet+FIS & \textbf{88.38}$\pm$0.11 & \textbf{85.08}$\pm$1.21 & \textbf{88.38}$\pm$0.49 & \textbf{84.89}$\pm$1.30 & \textbf{88.11}$\pm$0.10 & \textbf{1.72}$\pm$0.56 & \textbf{3.65}$\pm$1.19 & \textbf{27.63}$\pm$2.41 \\ 
        \midrule

        \multirow{3}{*}{\textbf{DR}} & 3D ResNet & 92.22$\pm$0.53 & 81.97$\pm$1.78 & 95.94$\pm$3.83 & 85.13$\pm$1.63 & 93.91$\pm$0.94 & 5.07$\pm$1.74 & 11.71$\pm$3.99 & 6.03$\pm$3.30 \\
        & 3D ResNet+Adv & 92.24$\pm$0.31 & 84.85$\pm$1.31 & 95.13$\pm$1.73 & 87.49$\pm$1.26 & 93.26$\pm$0.59 & 3.52$\pm$0.39 & 8.31$\pm$0.87 & 7.19$\pm$2.46 \\
        & 3D ResNet+FIS & \textbf{93.37}$\pm$0.25 & \textbf{85.78}$\pm$0.96 & \textbf{94.70}$\pm$0.68 & \textbf{87.29}$\pm$0.41 & \textbf{94.81}$\pm$0.06 & \textbf{3.46}$\pm$0.40 & \textbf{8.06}$\pm$1.10 & \textbf{5.77}$\pm$2.24 \\ \midrule

        \multirow{3}{*}{\textbf{Glaucoma}} & 3D ResNet & 86.49$\pm$0.19 & 81.60$\pm$0.34 & 88.81$\pm$0.48 & 82.90$\pm$0.52 & 86.57$\pm$0.10 & 2.87$\pm$0.31 & 6.85$\pm$0.20 & 4.47$\pm$2.29 \\
        & 3D ResNet+Adv & 86.21$\pm$0.16 & 79.29$\pm$0.86 & 89.37$\pm$0.33 & 81.14$\pm$1.03 & 86.70$\pm$0.17 & 3.96$\pm$0.37 & 9.53$\pm$0.82  & 6.29$\pm$2.79 \\
        & 3D ResNet+FIS &\textbf{88.43}$\pm$0.21 & \textbf{82.61}$\pm$0.17 & \textbf{90.65}$\pm$0.17 & \textbf{83.86}$\pm$0.12 & \textbf{88.70}$\pm$0.14 & \textbf{2.81}$\pm$0.27 & \textbf{6.78}$\pm$0.27 & \textbf{4.23}$\pm$1.36 \\ 
        
		\bottomrule	
	\end{tabular}}
\label{tbl:oct_race_fis}
\end{table*}

\begin{table*}[!t]
	\centering
	\caption{
        Comparison of the proposed FIS method against other fairness baselines trained on \textbf{SLO fundus images} and evaluated on the protected attribute of \textbf{gender}.
	}
	\adjustbox{width=1\textwidth}{
	\begin{tabular}{C{10ex} L{18ex} C{12ex} C{12ex} C{12ex} C{12ex} C{12ex} C{12ex} C{12ex} }
		\toprule
          \multirow{2}{*}{\textbf{Disease}} & \multirow{2}{*}{\textbf{Method}} & \textbf{Overall} & \textbf{\multirow{2}{*}{\textbf{ES-AUC$\uparrow$}}} & \textbf{Female} & \textbf{Male} & \textbf{Mean} & \textbf{Max} & \textbf{\multirow{2}{*}{\textbf{DEOdds$\downarrow$}}} \\
	 & & \textbf{AUC$\uparrow$} &  & \textbf{AUC$\uparrow$} & \textbf{AUC$\uparrow$} & \textbf{PSD$\downarrow$} & \textbf{PSD$\downarrow$} & \\
		\cmidrule(lr){1-1} \cmidrule(lr){2-2} \cmidrule(lr){3-3} \cmidrule(lr){4-4} \cmidrule(lr){5-6}  \cmidrule(lr){7-8} \cmidrule(lr){9-9}

        \multirow{4}{*}{\textbf{AMD}} & ViT & 82.32$\pm$0.17 & 80.30$\pm$0.26 & 83.23$\pm$0.25 & 80.77$\pm$0.22 & 1.52$\pm$0.22 & 3.04$\pm$0.44 & 10.95$\pm$3.04  \\
        & ViT+Adv & 81.97$\pm$0.21 & 79.67$\pm$0.50 & 83.00$\pm$0.10 & 80.17$\pm$0.40 & 1.76$\pm$0.24 & 3.51$\pm$0.47 & 8.87$\pm$2.83 \\
        & ViT+GroupDRO & 81.93$\pm$0.05 & 80.03$\pm$0.05 & 82.80$\pm$0.10 & 80.43$\pm$0.06 & 1.45$\pm$0.09 & 2.89$\pm$0.17 & 15.25$\pm$5.99  \\
        & ViT+FSCL & 81.96$\pm$0.03 & 80.04$\pm$0.07 & 82.88$\pm$0.12 & 80.45$\pm$0.07 & 1.46$\pm$0.05 & 2.92$\pm$0.13 & \textbf{7.14}$\pm$3.65 \\
        & ViT+FIS & \textbf{82.55}$\pm$0.33 & \textbf{80.46}$\pm$0.40 & \textbf{83.67}$\pm$0.36 & \textbf{80.93}$\pm$0.40 & \textbf{1.41}$\pm$0.24 & \textbf{2.66}$\pm$0.19 & 9.00$\pm$1.48 \\ \midrule

        \multirow{4}{*}{\textbf{DR}} & ViT & 85.22$\pm$0.21 & 84.58$\pm$0.64 & 84.84$\pm$0.54 & 85.43$\pm$0.44 & 0.49$\pm$0.42 & 0.87$\pm$0.83 & 5.48$\pm$4.40  \\
        & ViT+Adv & 84.35$\pm$0.14 & 83.37$\pm$0.55 & 84.22$\pm$0.72 & 85.39$\pm$0.58 & 0.69$\pm$0.55 & 1.39$\pm$0.95 & 6.75$\pm$3.26   \\
        & ViT+GroupDRO & 84.89$\pm$0.35 & 84.45$\pm$0.36 & 84.92$\pm$0.50 & 85.38$\pm$0.61 & 0.27$\pm$0.17 & 0.54$\pm$0.63 & 3.99$\pm$1.87  \\
        & ViT+FSCL & 85.32$\pm$0.64 & 84.56$\pm$0.73 & 85.96$\pm$0.75 & 85.06$\pm$0.40 & 0.53$\pm$0.85 & 1.05$\pm$0.77 & 5.26$\pm$2.68  \\
        & ViT+FIS & \textbf{85.69}$\pm$0.28 & \textbf{85.43}$\pm$0.82 & \textbf{85.86}$\pm$0.11 & \textbf{85.46}$\pm$0.60 & \textbf{0.25}$\pm$0.19 & \textbf{0.42}$\pm$0.28 & \textbf{2.36}$\pm$1.01  \\ \midrule

        \multirow{4}{*}{\textbf{Glaucoma}} & ViT & 79.73$\pm$0.18 & 77.81$\pm$0.29 & 78.63$\pm$0.18 & 81.11$\pm$0.38 & 1.59$\pm$0.29 & 3.13$\pm$0.58 & 2.66$\pm$0.78  \\
        & ViT+Adv  & 79.73$\pm$0.21 & 77.83$\pm$0.17 & 78.67$\pm$0.15 & 81.13$\pm$0.37 & 2.73$\pm$1.84 & 5.46$\pm$3.68 & \textbf{2.43}$\pm$0.23  \\
        & ViT+GroupDRO & 79.82$\pm$0.19 & 77.65$\pm$0.22 & 78.55$\pm$0.37 & 81.45$\pm$0.25 & 1.77$\pm$0.76 & 3.53$\pm$1.54 & 2.46$\pm$0.56   \\
        & ViT+FSCL & 80.47$\pm$0.54 & 77.88$\pm$0.57 & 79.12$\pm$0.88 & \textbf{82.44}$\pm$0.37 & 2.06$\pm$1.33 & 4.12$\pm$1.79 & 4.65$\pm$1.54 \\
        & ViT+FIS & \textbf{80.69}$\pm$0.27 & \textbf{78.85}$\pm$0.35 & \textbf{79.62}$\pm$0.26 & 81.94$\pm$0.38 & \textbf{1.46}$\pm$0.27 & \textbf{2.84}$\pm$0.53 & 2.54$\pm$0.37 \\ 
        
		\bottomrule	
	\end{tabular}}
\label{tbl:slo_gender_fis}
\end{table*}

\begin{table*}[!t]
	\centering
	\caption{
        Comparison of the proposed FIS method against other fairness baselines trained on \textbf{OCT B-Scans} and evaluated on the protected attribute of \textbf{gender}.
	}
	\adjustbox{width=1\textwidth}{
	\begin{tabular}{C{10ex} L{18ex} C{12ex} C{12ex} C{12ex} C{12ex} C{12ex} C{12ex} C{12ex} }
		\toprule
          \multirow{2}{*}{\textbf{Disease}} & \multirow{2}{*}{\textbf{Method}} & \textbf{Overall} & \textbf{\multirow{2}{*}{\textbf{ES-AUC$\uparrow$}}} & \textbf{Female} & \textbf{Male} & \textbf{Mean} & \textbf{Max} & \textbf{\multirow{2}{*}{\textbf{DEOdds$\downarrow$}}} \\
	 & & \textbf{AUC$\uparrow$} &  & \textbf{AUC$\uparrow$} & \textbf{AUC$\uparrow$} & \textbf{PSD$\downarrow$} & \textbf{PSD$\downarrow$} & \\
		\cmidrule(lr){1-1} \cmidrule(lr){2-2} \cmidrule(lr){3-3} \cmidrule(lr){4-4} \cmidrule(lr){5-6}  \cmidrule(lr){7-8} \cmidrule(lr){9-9}

        \multirow{3}{*}{\textbf{AMD}} & 3D ResNet & 87.63$\pm$0.29 & 86.39$\pm$0.73 & 88.14$\pm$0.13 & 86.70$\pm$0.62 & 0.82$\pm$0.32 & 1.64$\pm$0.64 & 6.44$\pm$0.48 \\
        & 3D ResNet+Adv & 86.88$\pm$0.38 & 86.85$\pm$0.85 & 86.89$\pm$0.16 & 86.85$\pm$0.72 & \textbf{0.18}$\pm$0.08 & \textbf{0.44}$\pm$0.14 & 4.08$\pm$1.22 \\
        & 3D ResNet+FIS & \textbf{88.34}$\pm$0.16 & \textbf{87.42}$\pm$0.25 & \textbf{88.70}$\pm$0.13 & \textbf{87.65}$\pm$0.20 & 0.59$\pm$0.16 & 1.19$\pm$0.31 & \textbf{4.98}$\pm$1.16 \\ \midrule

        \multirow{3}{*}{\textbf{DR}} & 3D ResNet & 92.22$\pm$0.53 & 91.05$\pm$0.89 & 92.70$\pm$0.49 & 91.42$\pm$0.57 & 0.69$\pm$0.15 & 1.38$\pm$0.30 & 2.61$\pm$2.25 \\
        & 3D ResNet+Adv & 93.20$\pm$0.71 & 92.43$\pm$0.57 & 93.46$\pm$0.66 & 92.63$\pm$0.97 & 0.45$\pm$0.26 & 0.90$\pm$0.43 & 1.98$\pm$0.77 \\
        & 3D ResNet+FIS & \textbf{93.41}$\pm$0.38 & \textbf{92.56}$\pm$0.18 & \textbf{93.91}$\pm$0.18 & \textbf{92.72}$\pm$0.58 & \textbf{0.39}$\pm$0.13 & \textbf{0.78}$\pm$0.27 & \textbf{1.71}$\pm$0.90 \\ \midrule

        \multirow{3}{*}{\textbf{Glaucoma}} & 3D ResNet & 86.49$\pm$0.19 & 83.53$\pm$0.71 & 84.93$\pm$0.60 & 88.47$\pm$0.39 & 2.05$\pm$0.55 & 4.10$\pm$1.10 & 8.84$\pm$1.44 \\
        & 3D ResNet+Adv & 86.62$\pm$0.50 & 83.35$\pm$0.49 & 84.89$\pm$0.35 & 88.81$\pm$0.73 & 2.26$\pm$0.30 & 4.53$\pm$0.60 & 8.75$\pm$0.65 \\
        & 3D ResNet+FIS & \textbf{88.68}$\pm$0.58 & \textbf{84.60}$\pm$0.21 & \textbf{86.48}$\pm$0.12 & \textbf{90.80}$\pm$0.14 & \textbf{1.81}$\pm$0.14 & \textbf{3.43}$\pm$0.32 & \textbf{7.23}$\pm$0.85 \\ 
        
		\bottomrule	
	\end{tabular}}
\label{tbl:oct_gender_fis}
\end{table*}

\begin{table*}[!t]
	\centering
	\caption{
        Comparison of the proposed FIS method against other fairness baselines trained on \textbf{SLO fundus images} and evaluated on the protected attribute of \textbf{ethnicity}.
	}
	\adjustbox{width=1\textwidth}{
	\begin{tabular}{C{10ex} L{18ex} C{12ex} C{12ex} C{12ex} C{12ex} C{12ex} C{12ex} C{12ex} }
		\toprule
          \multirow{2}{*}{\textbf{Disease}} & \multirow{2}{*}{\textbf{Method}} & \textbf{Overall} & \textbf{\multirow{2}{*}{\textbf{ES-AUC$\uparrow$}}} & \textbf{Non-Hisp} & \textbf{Hispanic} & \textbf{Mean} & \textbf{Max} & \textbf{\multirow{2}{*}{\textbf{DEOdds$\downarrow$}}} \\
	 & & \textbf{AUC$\uparrow$} &  & \textbf{AUC$\uparrow$} & \textbf{AUC$\uparrow$} & \textbf{PSD$\downarrow$} & \textbf{PSD$\downarrow$} & \\
		\cmidrule(lr){1-1} \cmidrule(lr){2-2} \cmidrule(lr){3-3} \cmidrule(lr){4-4} \cmidrule(lr){5-6}  \cmidrule(lr){7-8} \cmidrule(lr){9-9}

        \multirow{4}{*}{\textbf{AMD}} & ViT & 82.32$\pm$0.17 & 80.11$\pm$0.95 & 82.24$\pm$0.13 & 84.99$\pm$1.48 & 1.69$\pm$0.84 & 3.36$\pm$1.68 & 26.52$\pm$6.13  \\
        & ViT+Adv & 81.60$\pm$0.10 & 79.93$\pm$1.51 & 81.56$\pm$0.11 & 83.76$\pm$1.75 & \textbf{1.34}$\pm$1.14 & \textbf{2.70}$\pm$2.28 & \textbf{17.83}$\pm$4.04 \\
        & ViT+GroupDRO & 82.20$\pm$0.20 & 79.97$\pm$1.19 & 82.13$\pm$0.21 & 84.86$\pm$1.60 & 1.68$\pm$0.95 & 3.36$\pm$1.91 & 27.33$\pm$10.87  \\
        & ViT+FSCL & 82.41$\pm$0.17 & 79.05$\pm$1.28 & 82.37$\pm$0.19 & 85.35$\pm$1.38 & 2.59$\pm$1.57 & 5.17$\pm$1.77 & 35.63$\pm$8.99   \\
        & ViT+FIS & \textbf{82.69}$\pm$0.30 & \textbf{80.43}$\pm$0.71 & \textbf{82.57}$\pm$0.28 & \textbf{85.42}$\pm$1.54 & 1.39$\pm$0.76 & 2.93$\pm$1.52 & 24.56$\pm$8.02 \\ \midrule

        \multirow{4}{*}{\textbf{DR}} & ViT & 85.22$\pm$0.21 & 82.16$\pm$1.18 & 84.95$\pm$0.24 & 88.65$\pm$1.24 & 2.16$\pm$0.80 & 4.37$\pm$1.59 & 8.65$\pm$3.31  \\
        & ViT+Adv  & 85.37$\pm$0.75 & 82.39$\pm$1.20 & 85.16$\pm$0.57 & 88.77$\pm$0.93 & 2.11$\pm$1.04 & 4.23$\pm$1.44 & 6.33$\pm$2.74   \\
        & ViT+GroupDRO & 84.91$\pm$0.98 & 81.81$\pm$0.85 & 83.96$\pm$0.73 & 87.74$\pm$1.25 & 2.26$\pm$1.52 & 4.45$\pm$1.74 & 8.43$\pm$2.86  \\
        & ViT+FSCL & 85.67$\pm$1.38 & 82.31$\pm$0.75 & 84.83$\pm$0.60 & 88.92$\pm$0.94 & 2.39$\pm$0.97 & 4.77$\pm$1.33 & 5.98$\pm$1.97   \\
        & ViT+FIS & \textbf{85.80}$\pm$0.37 & \textbf{84.59}$\pm$1.30 & \textbf{85.66}$\pm$0.39 & \textbf{87.07}$\pm$1.42 & \textbf{0.81}$\pm$0.56 & \textbf{1.63}$\pm$0.93 & \textbf{4.07}$\pm$1.76 \\ \midrule

        \multirow{4}{*}{\textbf{Glaucoma}} & ViT & 79.73$\pm$0.18 & 73.89$\pm$0.94 & 80.00$\pm$0.15 & 72.18$\pm$1.28 & 4.88$\pm$0.79 & 9.83$\pm$1.58 & 10.23$\pm$1.62  \\
        & ViT+Adv & 80.06$\pm$0.45 & 74.63$\pm$1.40 & 80.33$\pm$0.40 & 73.00$\pm$1.81 & 4.57$\pm$0.91 & 9.15$\pm$1.84 & 9.62$\pm$2.34    \\
        & ViT+GroupDRO & 79.57$\pm$0.39 & 74.88$\pm$0.76 & 79.23$\pm$0.39 & 73.65$\pm$1.33 & \textbf{3.51}$\pm$1.62 & \textbf{7.01}$\pm$1.28 & 10.78$\pm$1.85    \\
        & ViT+FSCL & 80.16$\pm$0.37 & 75.31$\pm$0.97 & 80.55$\pm$0.89 & 74.11$\pm$0.99 & 4.02$\pm$0.76 & 8.03$\pm$2.13 & 8.62$\pm$1.57 \\
        & ViT+FIS & \textbf{80.58}$\pm$0.37 & \textbf{76.01}$\pm$1.14 & \textbf{80.86}$\pm$0.20 & \textbf{74.72}$\pm$1.50 & 3.78$\pm$0.84 & 7.55$\pm$1.69 & \textbf{7.39}$\pm$1.60 \\ 
		\bottomrule	
	\end{tabular}}
\label{tbl:slo_ethnicity_fis}
\end{table*}

\begin{table*}[!t]
	\centering
	\caption{
        Comparison of the proposed FIS method against other fairness baselines trained on \textbf{OCT B-Scans} and evaluated on the protected attribute of \textbf{ethnicity}.
	}
	\adjustbox{width=1\textwidth}{
	\begin{tabular}{C{10ex} L{18ex} C{12ex} C{12ex} C{12ex} C{12ex} C{12ex} C{12ex} C{12ex} }
		\toprule
          \multirow{2}{*}{\textbf{Disease}} & \multirow{2}{*}{\textbf{Method}} & \textbf{Overall} & \textbf{\multirow{2}{*}{\textbf{ES-AUC$\uparrow$}}} & \textbf{Non-Hisp} & \textbf{Hispanic} & \textbf{Mean} & \textbf{Max} & \textbf{\multirow{2}{*}{\textbf{DEOdds$\downarrow$}}} \\
	 & & \textbf{AUC$\uparrow$} &  & \textbf{AUC$\uparrow$} & \textbf{AUC$\uparrow$} & \textbf{PSD$\downarrow$} & \textbf{PSD$\downarrow$} & \\
		\cmidrule(lr){1-1} \cmidrule(lr){2-2} \cmidrule(lr){3-3} \cmidrule(lr){4-4} \cmidrule(lr){5-6}  \cmidrule(lr){7-8} \cmidrule(lr){9-9}

        \multirow{3}{*}{\textbf{AMD}} & 3D ResNet & 87.63$\pm$0.29 & 84.84$\pm$0.94 & 87.72$\pm$0.32 & 84.43$\pm$1.15 & 1.88$\pm$0.55 & 3.88$\pm$1.09 & 22.97$\pm$4.00 \\
        & 3D ResNet+Adv & 86.88$\pm$0.38 & 85.62$\pm$1.40 & 86.90$\pm$0.33 & 85.43$\pm$1.85 & \textbf{0.85}$\pm$0.47 & \textbf{1.74}$\pm$0.94 & 19.95$\pm$0.73 \\
        & 3D ResNet+FIS & \textbf{88.32}$\pm$0.22 & \textbf{85.91}$\pm$0.90 & \textbf{88.39}$\pm$0.07 & \textbf{85.56}$\pm$1.04 & 1.60$\pm$0.59 & 3.23$\pm$1.18 & \textbf{11.07}$\pm$0.93  \\ \midrule

        \multirow{3}{*}{\textbf{DR}} & 3D ResNet & 92.22$\pm$0.53 & 91.26$\pm$0.44 & 92.08$\pm$0.17 & 93.13$\pm$2.90 & 0.57$\pm$0.32 & 1.28$\pm$1.05 & 10.24$\pm$3.83  \\
        & 3D ResNet+Adv & 92.93$\pm$1.06 & 92.21$\pm$0.83 & 92.96$\pm$1.15 & 92.18$\pm$0.22 & 0.42$\pm$0.46 & 0.85$\pm$0.90 & 11.17$\pm$4.80 \\
        & 3D ResNet+FIS & \textbf{93.32}$\pm$0.25 & \textbf{93.06}$\pm$0.27 & \textbf{93.33}$\pm$0.16 & \textbf{93.67}$\pm$0.15 & \textbf{0.18}$\pm$0.07 & \textbf{0.38}$\pm$0.17 & \textbf{9.89}$\pm$3.95 \\ \midrule

        \multirow{3}{*}{\textbf{Glaucoma}} & 3D ResNet & 86.49$\pm$0.19 & 82.99$\pm$0.39 & 86.67$\pm$0.21 & 82.64$\pm$0.65 & 2.33$\pm$0.49 & 4.86$\pm$0.98 & 15.87$\pm$4.99 \\
        & 3D ResNet+Adv & 86.02$\pm$0.12 & 82.37$\pm$0.67 & 86.20$\pm$0.11 & 81.78$\pm$0.76 & 2.57$\pm$0.38 & 5.13$\pm$0.76 & 8.91$\pm$1.81 \\
        & 3D ResNet+FIS & \textbf{88.65}$\pm$0.34 & \textbf{83.64}$\pm$0.57 & \textbf{88.90}$\pm$0.16 & \textbf{83.89}$\pm$0.73 & \textbf{2.22}$\pm$0.62 & \textbf{4.52}$\pm$0.81 & \textbf{8.37}$\pm$2.43 \\ 
        
		\bottomrule	
	\end{tabular}}
\label{tbl:oct_ethnicity_fis}
\end{table*}

In this section, we present a detailed comparison of our proposed FIS method against 3 SOTA fairness methods including fair adversarial training (Adv)~\cite{beutel2017data}, GroupDRO~\cite{sagawa2019distributionally}, and fair contrastive loss~\cite{wang2022fairness} (FSCL). Specifically, Tables~\ref{tbl:slo_race_fis},~\ref{tbl:slo_gender_fis}, and~\ref{tbl:slo_ethnicity_fis} show results of models trained on SLO fundus images whereas Tables~\ref{tbl:oct_race_fis},~\ref{tbl:oct_gender_fis}, and~\ref{tbl:oct_ethnicity_fis} delineate results of models trained on OCT B-Scans from our Harvard-FairVision dataset.

\noindent \textbf{Race Fairness:}
Tables \ref{tbl:slo_race_fis} and \ref{tbl:oct_race_fis} present a comprehensive comparison of the proposed Fair Identity Scaling (FIS) method against other fairness baselines on SLO fundus images and OCT B-Scans, respectively, when evaluated on the protected attribute of race.
Comparing the performance between SLO fundus images and OCT B-Scans, it is evident that models trained on OCT B-Scans generally achieve higher Overall AUC and ES-AUC scores. For instance, in AMD detection, the 3D ResNet baseline achieves an Overall AUC of 87.63\% and an ES-AUC of 83.93\% on OCT B-Scans, while the ViT baseline achieves an Overall AUC of 82.32\% and an ES-AUC of 74.51\% on SLO fundus images. Similarly, for DR and Glaucoma detection, the 3D ResNet baseline on OCT B-Scans outperforms the ViT baseline on SLO fundus images, with the Overall AUC increasing from 85.22\% to 92.22\% and from 79.73\% to 86.49\%, and the ES-AUC increasing from 78.62\% to 81.97\% and from 74.23\% to 81.60\%, respectively. These results suggest that OCT B-Scans provide more discriminative features for achieving better overall performance and fairness in eye disease detection compared to SLO fundus images.

Regarding the performance improvement on the minority Black group, FIS consistently achieves higher ES-AUC and Group-wise AUC scores while reducing Mean PSD, Max PSD, and DEOdds compared to the baselines on both SLO fundus images and OCT B-Scans. On SLO fundus images, FIS improves the ES-AUC from 74.51\% (ViT) to 76.64\% for AMD, from 78.62\% to 78.96\% for DR, and from 74.23\% to 76.25\% for Glaucoma. Moreover, FIS enhances the Black AUC from 76.13\% to 78.04\% for AMD, from 79.69\% to 80.89\% for DR, and from 76.46\% to 77.66\% for Glaucoma. Furthermore, FIS reduces the Mean PSD from 2.72\% to 1.89\%, the Max PSD from 6.45\% to 4.57\%, and the DEOdds from 42.38\% to 35.47\% for AMD detection on SLO fundus images. Similar improvements in fairness metrics are observed for DR and Glaucoma detection on SLO fundus images.

On OCT B-Scans, FIS consistently outperforms the 3D ResNet baseline in terms of ES-AUC and Group-wise AUC scores while reducing Mean PSD, Max PSD, and DEOdds for all three eye diseases. For example, in AMD detection, FIS improves the ES-AUC from 83.93\% to 85.08\% and the Black AUC from 84.30\% to 84.89\%, while reducing the Mean PSD from 1.95\% to 1.72\%, the Max PSD from 4.64\% to 3.65\%, and the DEOdds from 28.51\% to 27.63\%. Similar trends are observed for DR and Glaucoma detection on OCT B-Scans, with FIS achieving higher ES-AUC and Black AUC scores while reducing Mean PSD, Max PSD, and DEOdds scores compared to the 3D ResNet baseline.

\noindent  \textbf{Gender Fairness:} 
%
Tables \ref{tbl:slo_gender_fis} and \ref{tbl:oct_gender_fis} present a comprehensive comparison of the proposed FIS against other fairness baselines on SLO fundus images and OCT B-Scans, respectively, when evaluated on the protected attribute of gender.
Comparing the performance between SLO fundus images and OCT B-Scans, it is evident that models trained on OCT B-Scans generally achieve higher Overall AUC and ES-AUC scores. For instance, in AMD detection, the 3D ResNet baseline achieves an Overall AUC of 87.63\% and an ES-AUC of 86.39\% on OCT B-Scans, while the ViT baseline achieves an Overall AUC of 82.32\% and an ES-AUC of 80.30\% on SLO fundus images. Similarly, for DR and Glaucoma detection, the 3D ResNet baseline on OCT B-Scans outperforms the ViT baseline on SLO fundus images, with the Overall AUC increasing from 85.22\% to 92.22\% and from 79.73\% to 86.49\%, and the ES-AUC increasing from 84.58\% to 91.05\% and from 77.81\% to 83.53\%, respectively. These results suggest that OCT B-Scans provide more discriminative features for achieving better overall performance and fairness in eye disease detection compared to SLO fundus images.

Regarding the performance improvement on the minority female group, FIS consistently achieves higher ES-AUC and Group-wise AUC scores while reducing Mean PSD, Max PSD, and DEOdds compared to the baselines on both SLO fundus images and OCT B-Scans. On SLO fundus images, FIS improves the ES-AUC from 80.30\% (ViT) to 80.46\% for AMD, from 84.58\% to 85.43\% for DR, and from 77.81\% to 78.85\% for Glaucoma. Moreover, FIS enhances the Female AUC from 83.23\% to 83.67\% for AMD, from 84.84\% to 85.86\% for DR, and from 78.63\% to 79.62\% for Glaucoma. Additionally, FIS reduces the Mean PSD from 1.52\% to 1.41\%, the Max PSD from 3.04\% to 2.66\%, and the DEOdds from 10.95\% to 9.00\% for AMD detection on SLO fundus images. Similar improvements in fairness metrics are observed for DR and Glaucoma detection on SLO fundus images.

On OCT B-Scans, FIS consistently outperforms the 3D ResNet baseline in terms of ES-AUC and Group-wise AUC scores while reducing Mean PSD, Max PSD, and DEOdds for all three eye diseases. For example, in AMD detection, FIS improves the ES-AUC from 86.39\% to 87.42\% and the Female AUC from 88.14\% to 88.70\%, while reducing the Mean PSD from 0.82\% to 0.59\%, the Max PSD from 1.64\% to 1.19\%, and the DEOdds from 6.44\% to 4.98\%. Similar trends are observed for DR and Glaucoma detection on OCT B-Scans, with FIS achieving higher ES-AUC and Female AUC scores while reducing Mean PSD, Max PSD, and DEOdds compared to the 3D ResNet baseline.


\noindent  \textbf{Ethnicity Fairness:}
Tables \ref{tbl:slo_ethnicity_fis} and \ref{tbl:oct_ethnicity_fis} present a comprehensive comparison of the proposed Fair Identity Scaling (FIS) method against other fairness baselines on SLO fundus images and OCT B-Scans, respectively, when evaluated on the protected attribute of ethnicity.
Comparing the performance between SLO fundus images and OCT B-Scans, it is evident that models trained on OCT B-Scans generally achieve higher Overall AUC and ES-AUC scores. For instance, in AMD detection, the 3D ResNet baseline achieves an Overall AUC of 87.63\% and an ES-AUC of 84.84\% on OCT B-Scans, while the ViT baseline achieves an Overall AUC of 82.32\% and an ES-AUC of 80.11\% on SLO fundus images. Similarly, for DR and Glaucoma detection, the 3D ResNet baseline on OCT B-Scans outperforms the ViT baseline on SLO fundus images, with the Overall AUC increasing from 85.22\% to 92.22\% and from 79.73\% to 86.49\%, and the ES-AUC increasing from 82.16\% to 91.26\% and from 73.89\% to 82.99\%, respectively. These results suggest that OCT B-Scans provide more discriminative features for achieving better overall performance and fairness in eye disease detection compared to SLO fundus images.

Regarding the performance improvement on the minority Hispanic group, FIS consistently achieves higher ES-AUC and Group-wise AUC scores while reducing Mean PSD, Max PSD, and DEOdds compared to the baselines on both SLO fundus images and OCT B-Scans. On SLO fundus images, FIS improves the ES-AUC from 80.11\% (ViT) to 80.43\% for AMD, from 82.16\% (ViT) to 84.59\% for DR, and from 73.89\% (ViT) to 76.01\% for Glaucoma. Moreover, FIS enhances the Hispanic AUC from 84.99\% (ViT) to 85.42\% for AMD, maintains it at 87.07\% for DR, and improves it from 72.18\% (ViT) to 74.72\% for Glaucoma. Additionally, FIS reduces the Mean PSD from 1.69\% (ViT) to 1.39\%, the Max PSD from 3.36\% (ViT) to 2.93\%, and the DEOdds from 26.52\% (ViT) to 24.56\% for AMD detection on SLO fundus images. Similar improvements in fairness metrics are observed for DR and Glaucoma detection on SLO fundus images.

On OCT B-Scans, FIS consistently outperforms the 3D ResNet baseline in terms of ES-AUC and Group-wise AUC scores while reducing Mean PSD, Max PSD, and DEOdds for all three eye diseases. For example, in AMD detection, FIS improves the ES-AUC from 84.84\% to 85.91\% and the Hispanic AUC from 84.43\% to 85.56\%, while reducing the Mean PSD from 1.88\% to 1.60\%, the Max PSD from 3.88\% to 3.23\%, and the DEOdds from 22.97\% to 11.07\%. Similar trends are observed for DR and Glaucoma detection on OCT B-Scans, with FIS achieving higher ES-AUC and Hispanic AUC scores while reducing Mean PSD, Max PSD, and DEOdds compared to the 3D ResNet baseline.

\noindent  \textbf{Summary:} To summarize, the proposed FIS method improves both overall performance as well as fairness on the three eye disease detection tasks across all three protected attributes of race, gender, and ethnicity. Moreover, FIS also outperforms the three SOTA fairness methods -- Adv, GroupDRO, and FSCL -- thereby alleviating the biases across various demographic subgroups.


\begin{figure}[!t]
	\centering
        \includegraphics[width=0.15\textwidth]{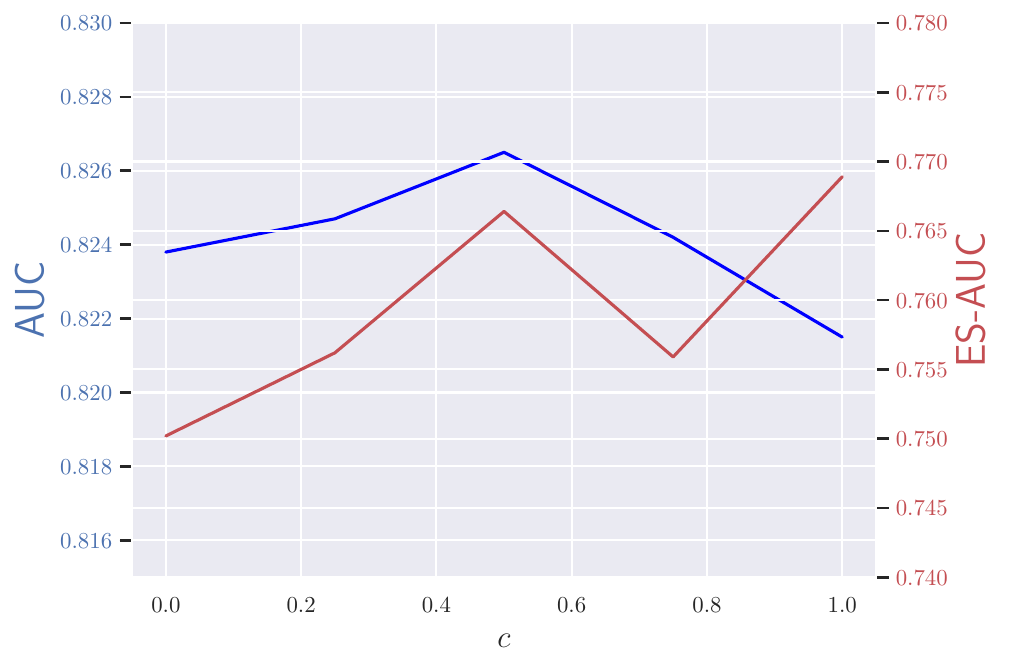} 
        \includegraphics[width=0.15\textwidth]{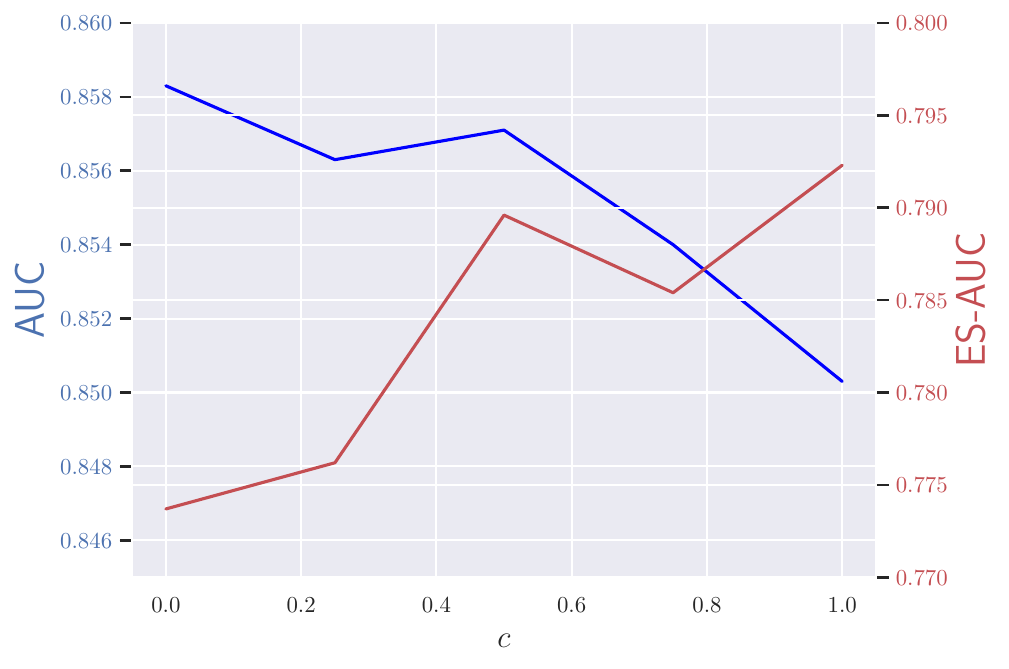} 
        \includegraphics[width=0.15\textwidth]{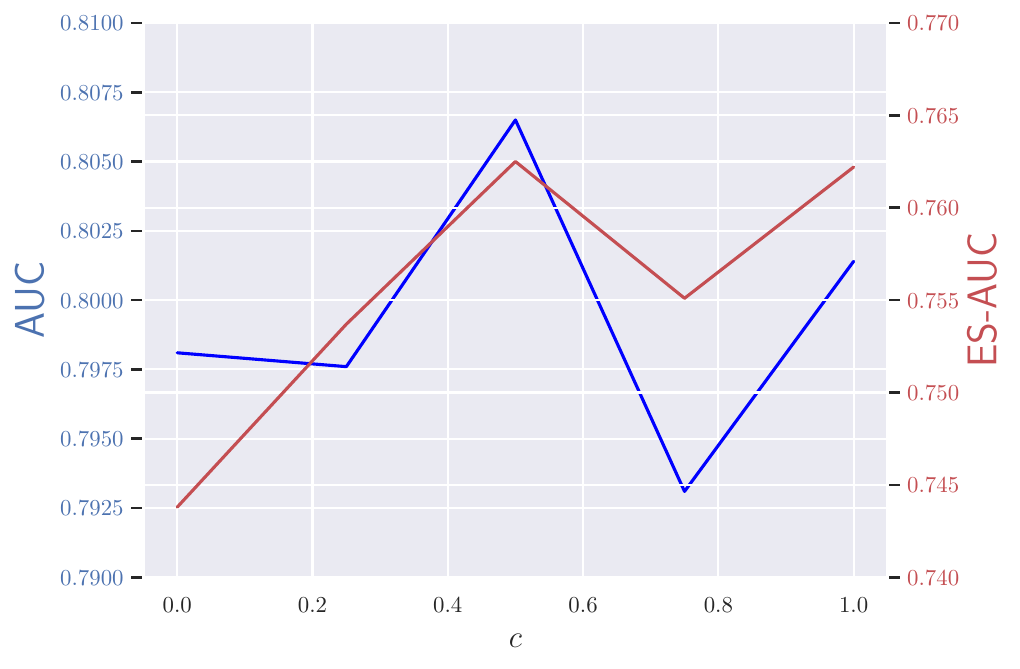}
	\caption{\label{fig:ablation_c}
    	Effects of the fusion weight $c$ on AUC and Mean PSD in AMD detection (left), DR detection (middle), and glaucoma detection (right). SLO fundus images are used for this analysis.
    	}
\end{figure}

\noindent  \textbf{Effects of Fusion Weight $c$:} The hyperparameter $c$ plays a central role in our proposed FIS. To gain insights into its impact on fairness learning, we conduct an ablation analysis with the race protected attribute, visualized in \textbf{Figure \ref{fig:ablation_c}}. The results reveal a notable trend: FIS, which combines both group-level and individual-level information, consistently outperforms scenarios involving only individual scaling ($c=0$) or group scaling ($c=1$). When $c$ is set to 0.5, we observe a balance between performance and demographic equity across all three disease types, demonstrating the effectiveness of $c=0.5$ in achieving a desirable trade-off.
	\section{Conclusion}

Equitable deep learning is profoundly important, particularly in healthcare, due to its direct impact on human health. While fairness in 2D medical imaging models has been studied, there is a lack of comprehensive evaluation regarding the fairness of 3D models. This gap is mainly due to the limited availability of large 3D datasets for fairness learning. Since 3D imaging data plays a crucial role in modern clinical practice and surpasses 2D imaging in clinical care quality, understanding fairness in 3D medical imaging models is crucial. To bridge this research gap, we conducted the first extensive study on fairness in 3D medical imaging diagnosis models across race, gender, and ethnicity attributes. Our study included 2D and 3D models, examining fairness across various architectures and eye diseases, uncovering notable biases across demographic subgroups. To mitigate these biases, we introduced the fair identity scaling (FIS) approach, utilizing group and individual scaling to enhance equity. Our results demonstrate that FIS not only improves overall performance but also fairness, surpassing several state-of-the-art fairness learning methods. Furthermore, we introduced Harvard-FairVision, a large-scale medical fairness dataset with 30,000 subjects, encompassing both 2D and 3D imaging data and six demographic attributes. Harvard-FairVision provides labeled data for three major eye disorders, making it a valuable resource for fairness learning in both 2D and 3D contexts.

	

	\bibliographystyle{IEEEtran}
	\bibliography{ref}
	
    
	
\end{document}